\documentclass[a4paper,fleqn]{cas-dc}

\usepackage[authoryear]{natbib}
\usepackage{algorithm}
\usepackage{algpseudocode}
\algrenewcommand\algorithmicrequire{\textbf{Input:}}
\algrenewcommand\algorithmicensure{\textbf{Output:}}
\usepackage{tabularx}
\usepackage{rotating}
\usepackage{float}
\usepackage{subcaption}

\usepackage{booktabs}
\usepackage{adjustbox}

\usepackage{graphicx}
\usepackage{array}
\usepackage{xcolor}

\newcommand{\figlabel}[1]{{\normalfont\normalsize\bfseries #1}}

\newcommand{\resulttablefont}{\fontsize{7.7}{9.5}\selectfont}

\def\tsc#1{\csdef{#1}{\textsc{\lowercase{#1}}\xspace}}
\tsc{WGM}
\tsc{QE}

\begin{document}
\let\WriteBookmarks\relax
\def\floatpagepagefraction{1}
\def\textpagefraction{.001}

\shorttitle{SimCortex v2: Joint Collision-Free Cortical Surface Reconstruction}    

\shortauthors{K. Moradkhani and S. Bouix}  

\title [mode = title]{SimCortex v2: Joint Cortical Surface Reconstruction with Near-Zero Collisions and Self-Intersections}  



%

\author[1]{Kaveh Moradkhani}[orcid=0009-0002-4372-073X]



\ead{kaveh.moradkhani.1@ens.etsmtl.ca}


\credit{Conceptualization, Methodology, Software, Formal analysis, Investigation, Data Curation, Writing - original Draft, Visualization}

\affiliation[1]{organization={École de technologie supérieure},
            city={Montréal},
            state={QC},
            country={Canada}}

\author[1]{Sylvain Bouix}[orcid=0000-0003-1326-6054]
\cormark[1]

\ead{sylvain.bouix@etsmtl.ca}
 

\credit{Conceptualization, Methodology, Resources, Writing - review \& editing, Supervision, Project administration, Funding acquisition}

\cortext[1]{Corresponding author}



\begin{abstract}
Reconstructing cortical WM and pial surfaces from structural magnetic resonance imaging (MRI) is a prerequisite for surface-based neuroanatomical analysis, yet remains challenging because the cortex is thin and tightly folded. 
Reconstruction methods can produce geometric artifacts such as mesh self-intersections and collisions between cortical surfaces, and although recent deep learning methods have reduced reconstruction time from hours to minutes, these artifacts persist. 
We propose SimCortex v2, a deep learning framework for simultaneous reconstruction of the left and right WM and pial surfaces from T1-weighted MRI. 
SimCortex v2 estimates topologically correct initial surfaces from a volumetric segmentation and refines all four jointly using multi-scale stationary velocity fields predicted by a ribbon-conditioned, U-Net-like network. 
We evaluated SimCortex v2 on 560 cases from 14 cohorts, thirteen of them unseen during training, spanning ages 6--89, healthy and clinical populations, and scanners from three vendors. 
SimCortex v2 matched the surface-distance accuracy of the strongest baseline (average symmetric surface distance $0.253$\,mm) while showing no detected inter-surface collision in $92.14\%$ of cases and the lowest self-intersection fraction ($0.044\%$) among learning-based methods, whereas every baseline produced at least one collision in every case. 
Source code, configuration files, pretrained weights, preprocessed data, and the exact evaluation splits are publicly released.
\end{abstract}


\begin{graphicalabstract}
\includegraphics[width=\textwidth]{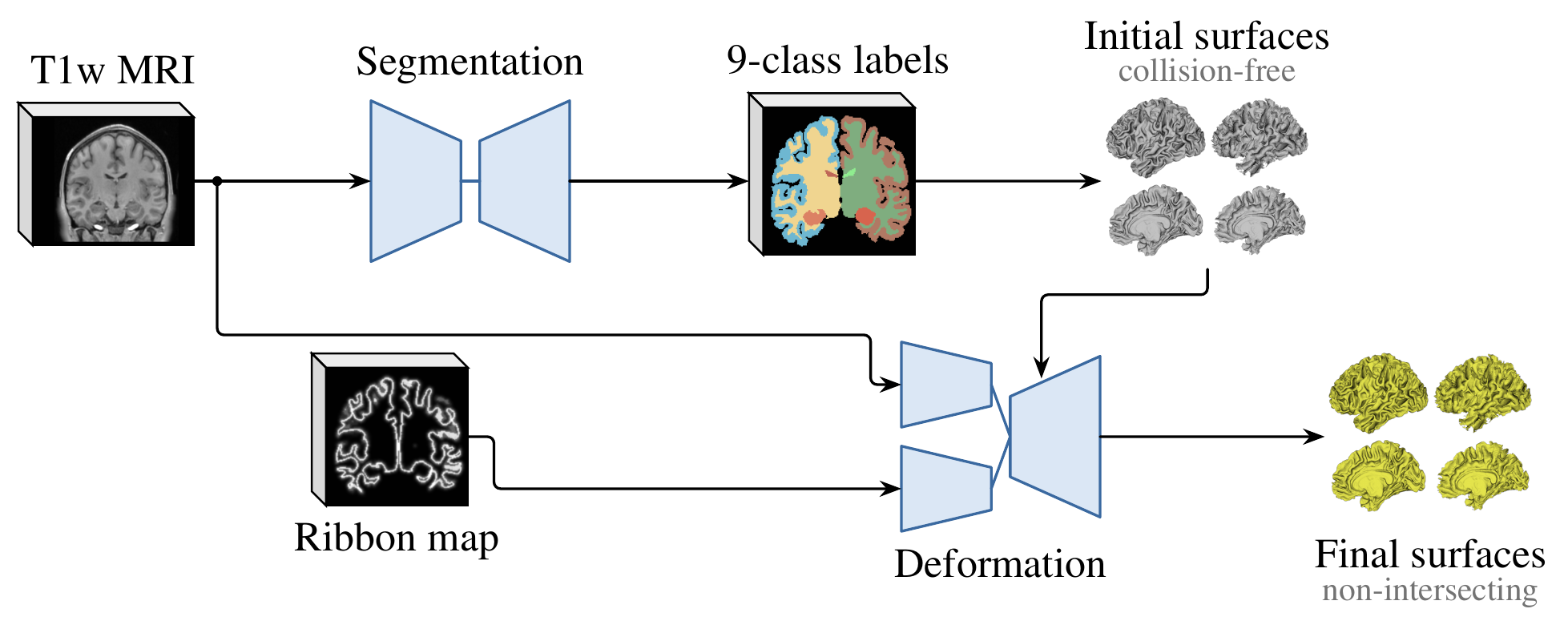}
\end{graphicalabstract}

\begin{highlights}
\item SimCortex v2 jointly reconstructs all four cortical surfaces from brain MRI
\item Ribbon-conditioned dual-encoder network with multiscale SVF deformation
\item No detected collision in 92\% of 560 cases from 14 multi-scanner datasets
\item Surface-distance accuracy comparable to the strongest evaluated baseline
\item Code, weights, processed data and evaluation splits released for reproduction
\end{highlights}



\maketitle

\section{Introduction}
\label{sec:intro}
 
Reconstructing the cerebral cortex from structural magnetic resonance imaging (MRI) is a central step in neuroimaging analysis. 
By explicitly modeling the inner white matter (WM) and outer pial boundaries, cortical surface representations support anatomically meaningful measurements of cortical morphometry, such as thickness, volume or surface area. 
These measurements are widely used to study brain development and aging \citep{shaw2016age} and to characterize structural alterations associated with neurological and psychiatric conditions, including Alzheimer's disease \citep{lerch2005focal,du2007different} and schizophrenia \citep{kuperberg2003regionally,rimol2012cortical}.
 
Cortical surface reconstruction remains challenging because the cortex is a thin, highly folded sheet whose tissue boundaries must be inferred from finite-resolution volumetric images. 
Partial-volume effects, limited spatial resolution, and local ambiguity in tissue contrast make the WM and pial interfaces difficult to localize precisely, particularly in narrow sulci and other high-curvature regions \citep{fischl2012freesurfer}. 
A reliable reconstruction method should therefore satisfy several requirements simultaneously:
\begin{itemize}
\item accurately recover fine-scale cortical geometry, including sulcal folds and high-curvature regions;
\item represent each cortical boundary as a topologically valid, genus-zero mesh with minimal self-intersections;
\item prevent anatomically implausible collisions or crossings between distinct cortical surfaces, including ipsilateral WM--pial interactions and inter-hemispheric surface interactions; and
\item remain computationally efficient for large neuroimaging cohorts and population-level analyses.
\end{itemize}
 
No existing approach satisfies all four requirements at once. 
Classical neuroimaging frameworks such as FreeSurfer \citep{fischl2012freesurfer,dale1999cortical} and BrainSuite \citep{shattuck2002brainsuite} are extensively validated and remain the reference pipelines for cortical surface reconstruction, but their multi-stage image- and geometry-processing workflows commonly require several hours per subject, and they do not explicitly model intersection avoidance or white--pial separation. 
Recent deep learning methods have reduced reconstruction time from hours to minutes, yet they trade the four requirements against one another: implicit formulations achieve sub-voxel boundary localization but require auxiliary mesh extraction and topology control, while explicit mesh-deformation methods preserve topology by construction but have addressed inter-surface consistency only for selected surface pairs. Consequently, anatomically implausible inter-surface interactions, including white--pial intersections and collisions involving surfaces from opposite hemispheres, remain common. 
We review these approaches in Section~\ref{sec:related_work}.
 
Our previous work, SimCortex v1 \citep{moradkhani2025simcortex}, targeted this gap by jointly estimating the left and right WM and pial boundaries from subject-specific, segmentation-driven initial meshes within a unified multiscale diffeomorphic deformation framework, reducing inter-surface collisions by more than 90\% relative to earlier explicit baselines. 
Surface placement accuracy, however, remained below the state of the art: errors were primarily associated with the quality of the WM surfaces in highly folded regions, where inaccuracies in the initial geometry propagated through the subsequent deformation and produced smoother, shallower sulcal reconstructions.
 
These observations motivate the present work. 
We introduce SimCortex v2, an extension of SimCortex v1 for simultaneous whole-brain cortical surface
reconstruction, designed to improve geometric fidelity in highly folded regions while preserving topologically valid surface representations and reducing anatomically implausible inter-surface contacts. 
Compared with SimCortex v1, SimCortex v2 introduces three main methodological changes:
 
\begin{itemize}
\item a hierarchical, subject-specific initialization strategy in which WM surfaces are generated directly from segmentation outputs with topology correction, and pial surfaces are subsequently derived from the WM signed distance fields through adaptive isolevel selection;
\item a deformation network conditioned on both the T1-weighted (T1w) MRI and a cortical ribbon probability map, providing additional spatial context for refining the initialized surfaces; and 
\item  a training objective which discourages insufficient separation between relevant surface pairs.
\end{itemize}
 
These components are integrated into a modular four-stage pipeline consisting of preprocessing, segmentation, hierarchical surface initialization, and diffeomorphic deformation. 
Together, this design improves the geometric quality of the initialized surfaces, incorporates ribbon-aware deformation cues, and explicitly regularizes inter-surface separation during training.

We evaluate SimCortex v2 on 560 cases drawn from 14 cohorts, spanning ages 6 to 89, healthy and clinical populations, and scanners from three vendors.

Every method, including the competing baselines, is scored in each subject's scanner right-anterior-superior (RAS) coordinate system against the same FreeSurfer reference surfaces using a single evaluation implementation.

To our knowledge, this is the most extensive and heterogeneous benchmark reported for deep learning cortical surface reconstruction; the compared methods were each originally validated on one or a few cohorts.
 
To make this benchmark reusable, we release the complete pipeline instead of the trained model alone. 
The source code is provided as a tagged release with installation instructions, configuration files, pretrained weights, and the commands used to produce every reported experiment. 
The processed data derivatives, the intermediate outputs of each pipeline stage, and the exact training, validation, and test splits together with the subject-selection lists are deposited under a persistent identifier (Section~\ref{sec:data_availability}). 

\section{Related work}
\label{sec:related_work}
 
\subsection{Traditional neuroimaging pipelines}
\label{sec:traditional_frameworks}
 
FreeSurfer \citep{fischl2012freesurfer,dale1999cortical} and BrainSuite \citep{shattuck2002brainsuite} reconstruct WM and pial surfaces from T1w MRI through multi-stage workflows combining tissue segmentation, mesh extraction, topology correction, and surface processing \citep{dale1999cortical,fischl1999cortical}. 
Because the mesh initially extracted from the WM segmentation may contain handles and holes, explicit
topology correction is applied before the corrected surface is expanded outward toward the pial boundary and subsequently inflated and registered to an atlas \citep{fischl2012freesurfer}. 
The Human Connectome Project (HCP) build on FreeSurfer surfaces with dedicated pre- and postprocessing and
surface registration for large-scale population analysis \citep{glasser2013minimal}. 
These frameworks provide the reference surfaces used to train and evaluate most learning-based methods, including the present work. 
They are not, however, guaranteed to be free of geometric inconsistencies: self-intersections and inter-surface collisions can occur in the resulting surfaces.
 
\subsection{Learning-based cortical surface reconstruction}
\label{sec:learning_based}
 
\paragraph{Implicit representations}
One class of learning-based methods represents cortical surfaces implicitly as continuous spatial functions rather than predicting triangulated meshes directly. 
DeepCSR \citep{cruz2021deepcsr} and Vox2Surf \citep{hong2021vox2surf} learn signed distance function representations from MRI volumes and recover the target surface as an isosurface of the predicted field, enabling sub-voxel localization without deforming a predefined template. 
Implicit fields do not, however, guarantee the spherical, genus-zero topology required for cortical analysis. These methods therefore require isosurface extraction, commonly via marching cubes \citep{lewiner2003efficient}, together with topology correction. 
Such auxiliary processing adds computational cost and may alter local geometry, particularly in narrow sulci where small geometric changes affect morphometric measurements \citep{bazin2007topology}.
 
\paragraph{Explicit mesh deformation}
A second class reconstructs cortical surfaces explicitly by deforming meshes with fixed connectivity toward the target anatomy, thereby avoiding isosurface extraction and inheriting the topology of the initialization.
Voxel2Mesh \citep{wickramasinghe2020voxel2mesh}, an influential precursor in medical image analysis, deforms a coarse spherical template using volumetric image features. 
CorticalFlow (CF) \citep{lebrat2021corticalflow} learns deformation fields that warp a predefined template toward either the WM or the pial boundary, and TopoFit \citep{hoopes2022topofit} follows a related template-fitting strategy for WM reconstruction by combining image and graph convolutions. 
Fixed connectivity alone, however, neither guarantees anatomical validity nor prevents self-intersections, and generic or over-smoothed templates may require large deformations to recover subject-specific sulcal geometry, limiting accuracy in narrow and high curvature regions.
 
To reduce this deformation burden, subsequent methods moved the initialization closer to subject-specific anatomy by exploiting the geometric relationship between the WM and pial boundaries. 
CorticalFlow++ (CFPP) \citep{santa2022corticalflow++} and PialNN \citep{ma2021pialnn} reconstruct the pial surface by deforming a predicted or reconstructed WM mesh, establishing vertex-wise correspondence between the two boundaries. 
CortexODE \citep{ma2022cortexode} extends this paradigm using neural ordinary differential equation flows, deforming a mesh extracted from a subject-specific WM segmentation toward the WM boundary and then expanding it toward the pial surface. 
Hybrid-CSR \citep{sun2023hybrid} instead combines both representations, deforming a coarse explicit mesh and then using oriented point estimates for differentiable Poisson surface reconstruction, with residual topological defects addressed by an optimization-based diffeomorphic correction stage. 
Anchoring reconstruction to subject-specific anatomy reduces the deformation required to recover complex folds, but because the surfaces are reconstructed sequentially, errors in the WM stage propagate to the pial surface.

\paragraph{Coupled and midthickness formulations}
Recognizing the limitations of sequential reconstruction, a further group of methods models the cortical boundaries in a coupled manner. 
Vox2Cortex (V2C) \citep{bongratz2022vox2cortex} combines convolutional image features with graph-based mesh refinement and exchanges geometric information between the inner and outer surfaces before every deformation block. 
V2C-Flow \citep{bongratz2024neural} parameterizes mesh deformation with neural ordinary differential equation flows and introduces virtual inter-surface edges between corresponding inner and outer vertices, which strengthen white--pial feature exchange and empirically reduce intersections; we return to the scope of this mechanism in Section~\ref{sec:shared_space}. 
A related line of work pursues stronger within-hemisphere coupling through midthickness-centered formulations, in which the WM and pial boundaries are evolved from, or displaced relative to, a common midthickness surface: CoCSR \citep{zheng2023coupled} within a coupled diffeomorphic framework, SurfNN \citep{zheng2023surfnn} and SurfNet \citep{zheng2025surfnet} through learned diffeomorphic mappings and half-thickness displacements, and SegCSR \citep{zheng2025segcsr} with additional weak supervision from cortical ribbon segmentation.
 
Coupling improves within-hemisphere consistency and reduces the risk of local white--pial intersections relative to purely sequential pipelines.
These mechanisms are nevertheless designed around corresponding boundaries within a hemisphere or between paired WM and pial surfaces. 
They do not impose explicit distance-based separation constraints across all relevant ipsilateral and contralateral surface pairs, so inter-surface separation is not explicitly optimized at the whole-brain level.
 
\subsection{Simultaneous deformation in a shared space and the remaining gap}
\label{sec:shared_space}
 
Several methods already reconstruct all four cortical surfaces in a single pass. 
V2C \citep{bongratz2022vox2cortex} and V2C-Flow \citep{bongratz2024neural} deform a whole-brain template spanning both hemispheres and both boundaries, and V2C-Flow models the WM and pial surfaces jointly with the explicit aim of avoiding intersections between them. 
What distinguishes SimCortex is therefore not the scope of the reconstruction but the space in which the deformation is parameterized. 
In V2C and V2C-Flow, displacements are predicted on the mesh itself, through graph convolutions and graph-parameterized neural ODEs respectively, with inter-surface consistency encouraged by feature exchange between corresponding inner and outer vertices. 
Smoothness of an individual vertex trajectory does not constrain the relative position of vertices belonging to different surfaces, so intersections remain possible and are reduced only empirically. 
CF \citep{lebrat2021corticalflow} and CFPP \citep{santa2022corticalflow++} do integrate velocity fields defined on the image grid, but apply a separate field to each surface in sequence, which likewise leaves cross-surface relationships unconstrained.
 
SimCortex v1 \citep{moradkhani2025simcortex} instead samples a single integrated velocity field, defined on the image grid, at the vertices of all four surfaces simultaneously. 
The four surfaces are thus transported by one diffeomorphism of the ambient space rather than by four separate mesh-level trajectories, so their relative arrangement is preserved by construction up to discretization. 
Optimizing the surfaces in this common deformation space yielded more than a 90\% relative reduction in inter-surface collision rates compared with earlier explicit baselines, including CFPP and V2C.

Its remaining limitations revealed an important gap. Although simultaneous deformation improved inter-surface consistency, geometric accuracy in highly folded regions remained sensitive to the quality of the initial WM surfaces, which could propagate through the subsequent deformation and yield smoother, shallower sulcal reconstructions than methods with stronger boundary-localization or surface-geometric priors \citep{moradkhani2025simcortex}. 
Simultaneous deformation in a shared space is therefore necessary but not sufficient for accurate cortical surface reconstruction: it must be combined with accurate subject-specific initialization, while retaining genus-zero topology and explicit control of inter-surface separation. 
SimCortex v2 addresses this by combining a revised hierarchical initialization with cortical ribbon conditioning and an explicit separation term during joint deformation.

\section{Methods}
\label{sec:methods}

\subsection{Overview of the framework}
\label{sec:method_overview}
 
SimCortex v2 reconstructs the four cortical surfaces \(\hat{\mathcal{S}} = \{\hat{S}_{L,\mathrm{wm}}, \hat{S}_{L,\mathrm{pial}},
\hat{S}_{R,\mathrm{wm}}, \hat{S}_{R,\mathrm{pial}}\}\) from a structural T1w MRI. 
Rather than estimating these boundaries independently, it formulates cortical surface reconstruction as a whole-brain problem in which all four surfaces are initialized in a common spatial frame and refined jointly.
 
\begin{figure*}[t]
\centering
\makebox[\textwidth][c]{\includegraphics[width=\textwidth]{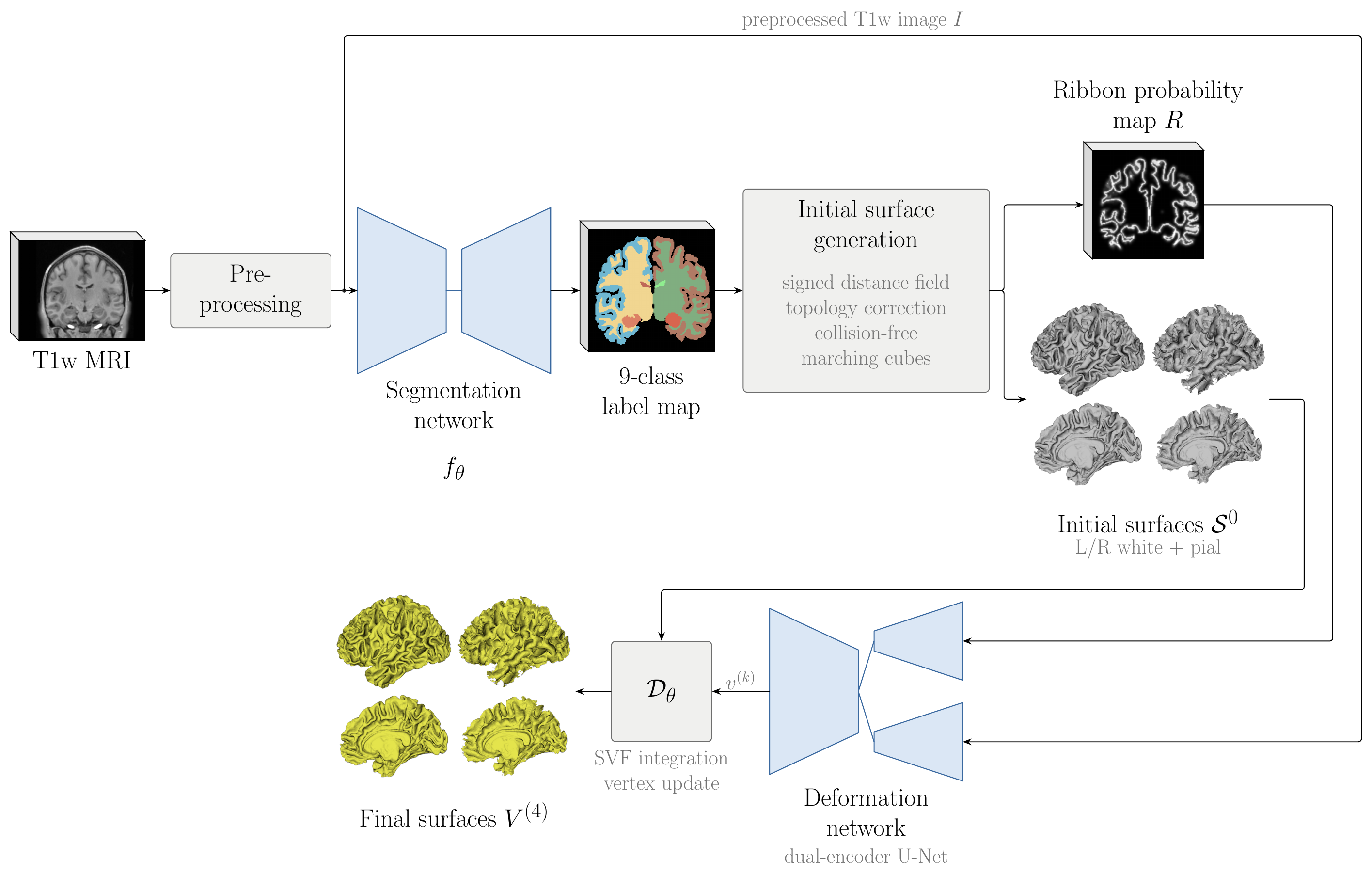}}
\caption{Overview of the SimCortex v2 pipeline. A T1w MRI is mapped to a common MNI152 space and passed to a 9-class segmentation network. The predicted anatomical labels are used to construct topology-corrected signed distance fields, from which subject-specific WM and pial initial surfaces and a cortical ribbon probability map are derived. A ribbon-conditioned deformation module then predicts multi-scale stationary velocity fields and jointly refines the four cortical surfaces in a shared deformation space.}
\label{fig:overview}
\end{figure*}

As illustrated in Fig.~\ref{fig:overview}, the pipeline has four stages: preprocessing to a common MNI152 space; volumetric segmentation into nine anatomical classes; hierarchical initialization of the four meshes and a cortical ribbon probability map, with topology correction and collision-aware isolevel selection; and ribbon-conditioned deformation, which predicts multi-scale stationary velocity fields (SVFs) and applies them jointly to the initialized meshes. 
Writing \(I\) for the preprocessed image, \(\hat{Y}\) for the predicted segmentation, \(\mathcal{S}^{0}\) for the initialized surfaces
and \(R\) for the ribbon map, 
\[
\hat{Y} = \mathcal{F}_{\phi}(I), \qquad
(\mathcal{S}^{0}, R) = \mathcal{G}(\hat{Y}), \qquad
\hat{\mathcal{S}} = \mathcal{D}_{\theta}(I, R, \mathcal{S}^{0}),
\]
where \(\mathcal{F}_{\phi}\) is the segmentation network, \(\mathcal{G}\) the deterministic initialization module, and \(\mathcal{D}_{\theta}\) the simultaneous deformation module. 
The remainder of this section describes the four stages in turn, then introduces the training objectives. 
Hyperparameters for every stage are presented in Table~\ref{tab:hyperparams}.

\subsection{Cortical surface reconstruction pipeline and network architecture}
\label{sec:inference}

\subsubsection{Preprocessing}
\label{sec:preprocessing}
 
For each subject, the input T1w MRI is corrected for intensity inhomogeneity using N4 bias-field correction \citep{tustison2010n4itk} and affinely aligned to the MNI152 T1 1\,mm template using ANTsPy \citep{avants2009advanced}. 
The native-to-MNI152 transform and its inverse are retained so that reconstructed surfaces can be exported back to the subject's native scanner space when needed.

\subsubsection{Volumetric segmentation}
\label{sec:segmentation}
 
\begin{figure*}[t]
\centering
\makebox[\textwidth][c]{\includegraphics[width=\textwidth]{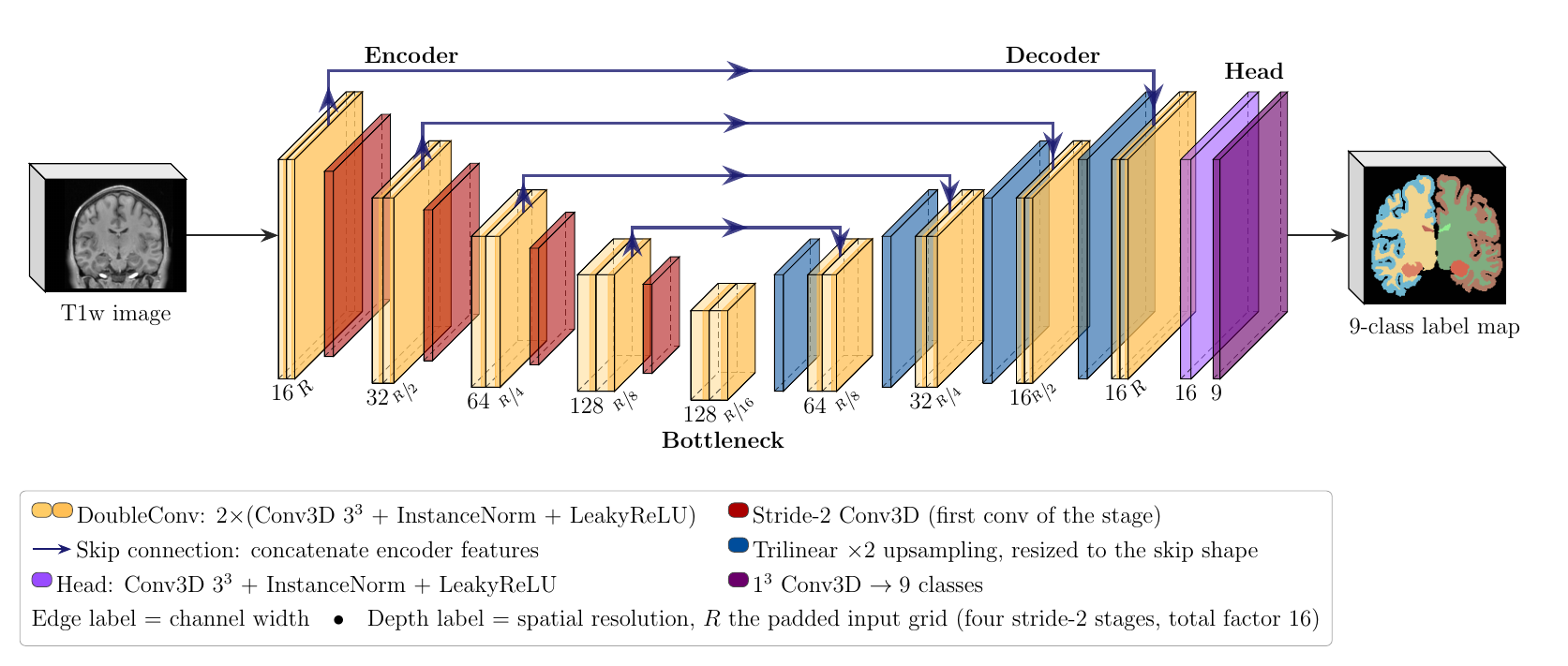}}
\caption{Volumetric segmentation network. The preprocessed T1w image is processed by a 3D U-Net encoder with four stride-2 downsampling stages, bottlenecked at $1/16$ resolution. Decoder stages upsample and fuse skip connections from each encoder level before a final classification head predicts voxel-wise logits over nine anatomical classes.}
\label{fig:segmentation_architecture}
\end{figure*}

The preprocessed image \(I\) is passed to a compact 3D U-Net (Fig.~\ref{fig:segmentation_architecture}) that predicts a 9-class anatomical label map \(\hat{Y}\), which provides the anatomical input to surface initialization.
 
The nine classes are defined from FreeSurfer \texttt{aparc+aseg} labels. 
The two inner-cerebrum/WM classes combine each hemisphere's cerebral WM with the deep gray-matter structures it encloses (thalamus, caudate, putamen, pallidum, accumbens and ventral diencephalon) together with adjacent choroid plexus, vessel and inferior lateral ventricle labels, so that the WM surface encloses them. 
The hippocampus--amygdala complex, the lateral ventricle and the cortical ribbon each form a separate hemisphere-specific class, and all remaining structures---including the third and fourth ventricles, brainstem
and extracerebral CSF---are treated as background.
The exact mappings between our labels and FreeSurfer labels are provided in Table~\ref{tab:seg9_mapping}. 

 Note that the network is trained on the 9-class labels, not the original FreeSurfer labels.

\subsubsection{Hierarchical surface initialization}
\label{sec:surface_initialization}

The initialization stage converts the predicted segmentation into four genus-zero meshes---the left and right WM and pial surfaces---together with a soft cortical ribbon map that conditions the subsequent deformation.
The stage is hierarchical: the WM surfaces are derived from the segmentation, the pial surfaces are derived from the WM signed distance fields (SDFs), and the ribbon map is derived from both. 
Its objective is to place the four surfaces as close to the target anatomy as the requirement that they remain pairwise disjoint allows, so that the deformation stage begins from a collision-free configuration. 
Parameters are listed in Table~\ref{tab:hyperparams} and the complete procedure is given in Algorithms~\ref{alg:wm_initialization} and~\ref{alg:pial_initialization}.

\paragraph{Hemisphere-specific WM masks and signed distance fields}
For each hemisphere, an inner seed mask is formed by grouping the inner cerebral label with the corresponding lateral ventricle label. 
Including the ventricles fills the interior of the mask, which prevents the WM level set from acquiring handles around the ventricular cavities. 
A midline gap is then opened before any surface is extracted: both masks are dilated by \(g\) voxels and every seed voxel in the overlap of the dilated masks is removed from both hemispheres, which is symmetric by construction and separates the seeds by approximately \(2g\) voxels. 
Each mask is reduced to its largest connected component and converted to an SDF, negative inside, which is
Gaussian-smoothed with \(\sigma_{\mathrm{sdf}}\) to suppress voxelization artifacts and corrected to genus zero by fast-marching topology correction \citep{bazin2007topology} with threshold \(T_{\mathrm{topo}}\).

\paragraph{Collision-aware WM level selection}
Since the SDF is negative inside, more negative isolevels lie further inside the WM boundary, so lowering the level contracts the surface away from the midline and the opposite hemisphere. 
Candidate levels are traversed inward from \(\tau_0\) in steps of \(\Delta_{\mathrm{wm}}\) down to \(\tau_{\min}\). 
At each level both hemispheres are extracted by marching cubes \citep{lewiner2003efficient}, Laplacian-smoothed, mapped to world coordinates, and tested for left--right intersection.
The first collision-free level, \(\tau_{\mathrm{free}}\), is retained. 
A single level is shared by both hemispheres so that the two surfaces remain symmetric with respect to the selection criterion.

By construction \(\tau_{\mathrm{free}}\) is the outermost collision-free level, and leaves no margin for the estimation of the initial pial surfaces.
A further inward inset of \(\delta_{\mathrm{wm}}\) voxels is therefore applied, clamped at \(\tau_{\min}\) and retained only if the resulting surfaces are still collision-free; otherwise the selection falls back to \(\tau_{\mathrm{free}}\). 
The selected level is denoted \(\tau_{\mathrm{wm}}\). 
If no collision-free candidate is found, the most inward level tested is returned and a warning is logged. This fallback did not occur in the 560 evaluation cases.

\paragraph{Re-centering and pial level selection}
Both fields are shifted so that the selected level  \(\tau_{\mathrm{wm}}\) becomes the zero level.
This sets the convention used by the pial search, in which positive values denote distance outward from the selected WM surfaces. 
We also reapply the topology correction procedure at the new zero level to guarantee genus zero of the WM surfaces.

Pial surfaces are obtained as outward level sets of this re-centered, re-corrected field. 
The pial initialization is therefore determined entirely by the WM field: no image evidence specific to the pial boundary enters at this stage, and the initial pial surface is a smooth, approximately constant-thickness offset of the WM surface. 
Recovering pial geometry is left to the deformation stage (Section \ref{sec:deformation}), which is one reason the ribbon map below is supplied to the network as an additional input.

A single symmetric offset, again shared by both hemispheres, is searched outward from \(\tau_{\mathrm{pial}}^{\min}\) to
\(\tau_{\mathrm{pial}}^{\max}\) in steps of \(\Delta_{\mathrm{pial}}\). 
Each candidate is tested first against the two ipsilateral WM surfaces and then, if those are clear, for left--right pial intersection. 
The valid  offsets form a single contiguous interval, so the meaning of a collision depends on where the search has reached. 
Before any valid offset has been found, a collision indicates that the search has not yet entered the
interval, and the walk outward continues. 
Once a valid offset has been found, a collision indicates that the search has just passed the far end of the collision-free interval, and the walk stops. 
The last valid offset is therefore retained, which places the pial surface as far outward as collision-freeness
permits. 
If no offset in the primary range is valid, a fallback search proceeds inward from below \(\tau_{\mathrm{pial}}^{\min}\) toward
\(\tau_{\mathrm{pial}}^{\mathrm{floor}}\) and accepts the first collision-free pair. 
Initialization fails if neither search succeeds. This did not occur in our experiments: all 560 evaluated cases completed initialization and passed the final checks for left–right WM, ipsilateral WM–pial, and left–right pial collisions.
Inter-surface collisions are assessed using triangle-level collision queries issued directly to the Flexible Collision Library (FCL)
\citep{pan2012fcl}.

\paragraph{Cortical ribbon probability map}
After the pial field is recentered in the same way, the bilateral cortical ribbon is defined as the region enclosed by the pial level set and outside the WM level set, unioned over hemispheres:
\begin{equation}
M_{\mathrm{rib}}
=
\bigcup_{h\in\{L,R\}}
\left\{
\mathbf{x}:
\phi_{h,\mathrm{pial}}(\mathbf{x}) \le 0
\ \wedge\
\phi_{h,\mathrm{wm}}(\mathbf{x}) > 0
\right\}.
\label{eq:ribbon_mask}
\end{equation}
An SDF is computed from \(M_{\mathrm{rib}}\) and smoothed with \(\sigma_{\mathrm{sdf}}\), here without largest-component filtering so that both hemispheres are retained, and converted to a probability map by
\begin{equation}
R(\mathbf{x})
=
\frac{1}{1+\exp\!\left(\beta\,\phi_{\mathrm{rib}}(\mathbf{x})\right)},
\label{eq:ribbon_prob}
\end{equation}
which is close to one inside the ribbon and close to zero outside, with \(\beta\) controlling the sharpness of the transition. 
The map \(R\) is provided to the deformation network as a second input channel alongside the T1w image.




\begin{algorithm}[t]
\footnotesize
\caption{White matter surface initialization.
\(\mathcal{E}(\phi,\tau)\) extracts a mesh at level \(\tau\) by marching
cubes, applies \(n_{\mathrm{smooth}}\) Laplacian iterations, and maps
its vertices to world coordinates with \(A\).
Extracted meshes are cached by hemisphere and level.
\(\textsc{Disjoint}(\tau)\) is true when
\(\mathcal{E}(\phi^{\mathrm{topo}}_{L,\mathrm{wm}},\tau)\) and
\(\mathcal{E}(\phi^{\mathrm{topo}}_{R,\mathrm{wm}},\tau)\)
do not intersect.
Parameters: Table~\ref{tab:hyperparams}.}
\label{alg:wm_initialization}
\begin{algorithmic}[1]

\Require Predicted segmentation \(\hat{Y}\), image affine \(A\)
\Ensure WM surfaces \(S^{0}_{h,\mathrm{wm}}\), fields
\(\phi_{h,\mathrm{wm}}\), \(\phi^{\mathrm{topo}}_{h,\mathrm{pial}}\)

\State \(M_L \gets [\hat{Y}\in\{1,7\}]\);\quad
       \(M_R \gets [\hat{Y}\in\{2,8\}]\)
\State \(O \gets \mathrm{dilate}(M_L,g)
             \cap \mathrm{dilate}(M_R,g)\)
\Comment{midline separation}
\State \(M_h \gets M_h \setminus O\) for \(h\in\{L,R\}\)
\For{\(h\in\{L,R\}\)}
    \State \(M_h \gets\) largest connected component of \(M_h\)
    \State \(\phi^{\mathrm{topo}}_{h,\mathrm{wm}} \gets
    \mathcal{T}_{\mathrm{topo}}\!\left(
    G_{\sigma_{\mathrm{sdf}}}*\mathrm{sdf}(M_h);
    T_{\mathrm{topo}}\right)\)
\EndFor

\Statex \emph{Select the outermost tested collision-free level,
then apply a collision-free inward inset if possible.}
\State \(\tau_{\mathrm{free}}\gets\textsc{None}\)
\State \(\tau_{\mathrm{last}}\gets\tau_0\)
\For{\(\tau=\tau_0,\ \tau_0+\Delta_{\mathrm{wm}},\ \ldots\),
      while \(\tau\geq\tau_{\min}\)}
    \State \(\tau_{\mathrm{last}}\gets\tau\)
    \If{\(\textsc{Disjoint}(\tau)\)}
        \State \(\tau_{\mathrm{free}}\gets\tau\);
               \textbf{break}
    \EndIf
\EndFor
\If{\(\tau_{\mathrm{free}}=\textsc{None}\)}
    \State \(\tau_{\mathrm{wm}}\gets\tau_{\mathrm{last}}\)
    \State issue a warning
    \Comment{not observed in the 560 evaluation cases}
\Else
    \State \(\tau_{\mathrm{inset}}\gets
    \max(\tau_{\min},
         \tau_{\mathrm{free}}-\delta_{\mathrm{wm}})\)
    \If{\(\textsc{Disjoint}(\tau_{\mathrm{inset}})\)}
        \State \(\tau_{\mathrm{wm}}\gets\tau_{\mathrm{inset}}\)
    \Else
        \State \(\tau_{\mathrm{wm}}\gets\tau_{\mathrm{free}}\)
    \EndIf
\EndIf

\For{\(h\in\{L,R\}\)}
    \State \(S^{0}_{h,\mathrm{wm}}\gets
    \mathcal{E}(\phi^{\mathrm{topo}}_{h,\mathrm{wm}},
                \tau_{\mathrm{wm}})\)
    \Comment{reuse cached mesh}
    \State \(\phi_{h,\mathrm{wm}}\gets
    \phi^{\mathrm{topo}}_{h,\mathrm{wm}}
    -\tau_{\mathrm{wm}}\)
    \State \(\phi^{\mathrm{topo}}_{h,\mathrm{pial}}\gets
    \mathcal{T}_{\mathrm{topo}}\!\left(
    \phi_{h,\mathrm{wm}};T_{\mathrm{topo}}\right)\)
\EndFor

\end{algorithmic}
\end{algorithm}

\begin{algorithm}[t]
\footnotesize
\caption{Pial offset selection and cortical ribbon map.
$\textsc{Check}(\tau)$ extracts or retrieves cached pial meshes
$P_h(\tau)=\mathcal{E}(\phi^{\mathrm{topo}}_{h,\mathrm{pial}},\tau)$.
It checks $P_L$ against $S^0_{L,\mathrm{wm}}$ and $P_R$ against
$S^0_{R,\mathrm{wm}}$ first. Only if both pairs are collision-free
does it check $P_L$ against $P_R$.
It returns \textsc{WM-Collision}, \textsc{LR-Collision}, or
\textsc{Valid}, and supplies the candidate meshes when valid.
Parameters: Table~\ref{tab:hyperparams}.}
\label{alg:pial_initialization}
\begin{algorithmic}[1]

\Require $S^{0}_{h,\mathrm{wm}}$, $\phi_{h,\mathrm{wm}}$,
$\phi^{\mathrm{topo}}_{h,\mathrm{pial}}$
\Ensure Initial surfaces $\mathcal{S}^{0}$, ribbon map $R$

\State $\tau_{\mathrm{pial}} \gets \textsc{None}$
\For{$\tau = \tau_{\mathrm{pial}}^{\min},\ \dots,\
\tau_{\mathrm{pial}}^{\max}$ \textbf{step} $\Delta_{\mathrm{pial}}$}
    \State $(s,P_L,P_R) \gets \textsc{Check}(\tau)$
    \If{$s=\textsc{Valid}$}
        \State $(\tau_{\mathrm{pial}},S^0_{L,\mathrm{pial}},
        S^0_{R,\mathrm{pial}}) \gets (\tau,P_L,P_R)$
    \ElsIf{$s=\textsc{LR-Collision}$}
        \State \textbf{break}
    \ElsIf{$\tau_{\mathrm{pial}}\neq\textsc{None}$}
        \State \textbf{break}
        \Comment{WM collision after a valid offset}
    \EndIf
\EndFor

\If{$\tau_{\mathrm{pial}}=\textsc{None}$}
    \For{$\tau = \tau_{\mathrm{pial}}^{\min}
    -\Delta_{\mathrm{pial}},\ \dots,\
    \tau_{\mathrm{pial}}^{\mathrm{floor}}$
    \textbf{step} $-\Delta_{\mathrm{pial}}$}
        \State $(s,P_L,P_R) \gets \textsc{Check}(\tau)$
        \If{$s=\textsc{Valid}$}
            \State $(\tau_{\mathrm{pial}},S^0_{L,\mathrm{pial}},
            S^0_{R,\mathrm{pial}}) \gets (\tau,P_L,P_R)$
            \State \textbf{break}
        \EndIf
    \EndFor
    \If{$\tau_{\mathrm{pial}}=\textsc{None}$}
        \State \textbf{raise} initialization failure
    \EndIf
\EndIf

\For{$h\in\{L,R\}$}
    \State $\phi_{h,\mathrm{pial}}\gets
    \phi^{\mathrm{topo}}_{h,\mathrm{pial}}-\tau_{\mathrm{pial}}$
\EndFor

\State $M_{\mathrm{rib}}\gets$ Eq.~\eqref{eq:ribbon_mask}
\If{$M_{\mathrm{rib}}=\varnothing$}
    \State $\phi_{\mathrm{rib}}\gets\mathbf{1}$;
    $R\gets\mathbf{0}$
\Else
    \State \textbf{try}:
    \State $\phi_{\mathrm{rib}}\gets
    G_{\sigma_{\mathrm{sdf}}}*\mathrm{sdf}(M_{\mathrm{rib}})$
    \Comment{retain all components}
    \State $R\gets$ Eq.~\eqref{eq:ribbon_prob}
    \State \textbf{on} failure:
    \State $\phi_{\mathrm{rib}}\gets\mathbf{1}$;
    $R\gets\mathbf{0}$
\EndIf

\State \Return $\mathcal{S}^{0}=
\{S^{0}_{h,\mathrm{wm}},S^{0}_{h,\mathrm{pial}}\}_{h\in\{L,R\}}$, $R$

\end{algorithmic}
\end{algorithm}

\subsubsection{Ribbon-conditioned deformation}
\label{sec:deformation}
 
The final stage jointly refines the initialized surfaces toward the anatomical WM and pial interfaces. 
It predicts volumetric SVFs from image and ribbon-derived cues, integrates them into displacement fields, and applies the result to the surface vertices in a shared volumetric frame. 
Mesh connectivity is fixed throughout, so the combinatorial topology of the initialized meshes is preserved by construction.
 
\paragraph{Network input and architecture}
\label{sec:def_arch}
 
The volumetric input is the two-channel tensor \([\,I,R\,]\), with the T1w image normalized by the mean and standard deviation of its nonzero voxels and the ribbon map clipped to \([0,1]\). 
The vertices of the four initialized surfaces are represented in the same voxel coordinate system and concatenated into a single matrix \(V^{(0)}\in\mathbb{R}^{N_v\times3}\) in the fixed order \((L,\mathrm{pial})\), \((L,\mathrm{wm})\), \((R,\mathrm{pial})\),
\((R,\mathrm{wm})\). 
Face connectivity and the per-surface vertex ranges are stored separately so that the deformed vertex tensor can be split back into four meshes.
 
\begin{figure*}[t]
\centering
\makebox[\textwidth][c]{\includegraphics[width=\textwidth]{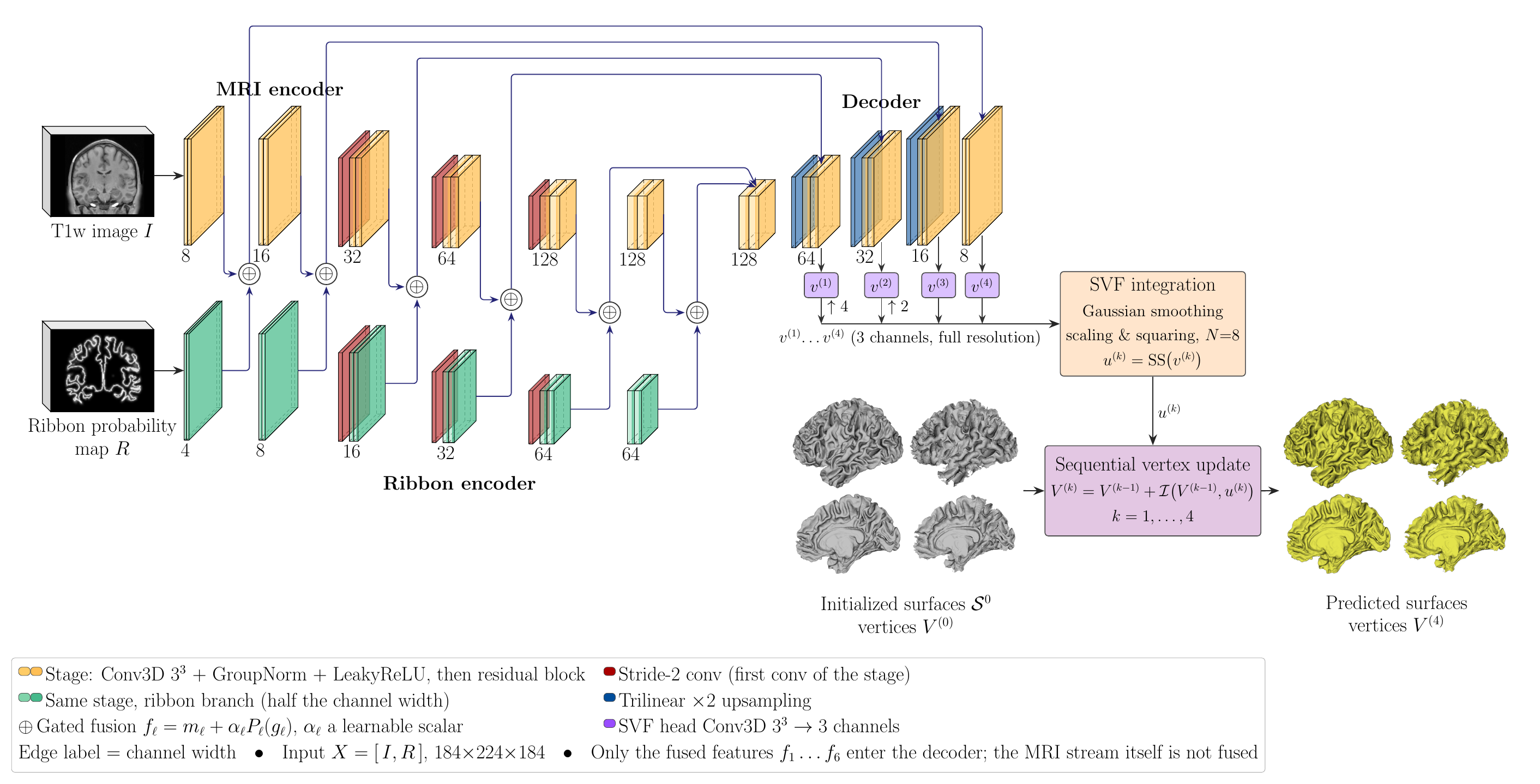}}
\caption{Ribbon-conditioned deformation architecture. 
The preprocessed T1w image and cortical ribbon probability map are processed by separate encoders.
Multi-scale features are fused through gated projections and decoded into stationary velocity fields, which are integrated by scaling and squaring and applied sequentially to the concatenated vertices of the four initialized cortical surfaces.}
\label{fig:deformation_architecture}
\end{figure*}

As shown in Fig.~\ref{fig:deformation_architecture}, the deformation network \(\mathcal{D}_{\theta}\) is a dual-encoder U-Net: one encoder processes the T1w image, a second, lighter encoder processes the ribbon map. 
Both use residual 3D convolutional stages. 
The first two stages operate at full resolution, the next three downsample to \(1/2\), \(1/4\) and \(1/8\), and a final stage refines at \(1/8\). 
At each stage \(\ell\), ribbon features are admitted into the MRI stream through a gated projection,
\begin{equation}
f_\ell = m_\ell + \operatorname{sigmoid}(a_\ell)\,P_\ell(g_\ell),
\label{eq:gated_fusion}
\end{equation}
where \(m_\ell\) and \(g_\ell\) are the MRI and ribbon features, \(P_\ell\) is a \(1\times1\times1\) projection and \(a_\ell\) is a learnable scalar gate.
Initializing \(a_\ell\) at \(-3\) opens the gate at roughly \(0.05\), so the ribbon branch starts with almost no influence and the network learns how much to admit. 
The fused features \(f_\ell\) serve as decoder skip features. 
The decoder restores resolution by trilinear interpolation and predicts four SVFs---one at 1/4, one at 1/2 and two at full resolution---which are upsampled to the full deformation grid before integration.

\paragraph{Integration and sequential surface update}
\label{sec:def_integration}
 
Each field is Gaussian-smoothed and integrated by scaling and squaring \citep{arsigny2006log} with \(N\) steps, yielding displacement fields \(u^{(1)},\ldots,u^{(4)}\). 
The four surfaces are then \textit{deformed together as a single concatenated vertex set}, the fields being applied sequentially in decoder order rather than summed:
\begin{equation}
V^{(k)} = V^{(k-1)} + \mathcal{I}\!\left(V^{(k-1)}, u^{(k)}\right),
\qquad k=1,\ldots,4,
\label{eq:sequential_update}
\end{equation}
where \(\mathcal{I}(V,u)\) trilinearly interpolates \(u\) at the vertex positions \(V\). 
Each field therefore acts on the vertices produced by the preceding update.
 
Two properties of this construction matter for what follows. 
First, each \(u^{(k)}\) is defined on the image grid and is sampled at the vertices of all four surfaces, so the composition of the four updates is a single deformation of the ambient space rather than four independent per-surface trajectories. 
A diffeomorphism of the ambient space is a bijection, so surfaces that are pairwise disjoint before deformation remain pairwise disjoint after it. 
In the discrete implementation this holds only approximately: scaling and squaring is exactly invertible in the continuous limit but not after \(N\) discrete steps, displacements are trilinearly interpolated, and two continuous surfaces separated by less than the local triangle scale can still yield intersecting piecewise-linear approximations.
Second, because connectivity is fixed, the genus of each surface cannot change. 
After the final update, \(V^{(4)}\) is split back into the four surfaces using the stored vertex ranges, retaining the original face connectivity.

\subsection{Training objectives}
\label{sec:training}
 
Supervision targets are derived from the FreeSurfer 7.4.1 outputs for each training subject and aligned to MNI152 space. 
For segmentation, the \texttt{aparc+aseg.mgz} and \texttt{filled.mgz} label volumes are resampled onto the MNI152 reference grid using nearest-neighbor interpolation and the subject-specific T1w-to-MNI152 affine registration (Section~\ref{sec:preprocessing}). 
The aligned \texttt{aparc+aseg} labels are mapped to nine classes (Table~\ref{tab:seg9_mapping}); \texttt{filled} is used to assign ambiguous labels 77 and 80 to the appropriate hemisphere.
For deformation, the FreeSurfer white matter and pial meshes of both hemispheres are transformed from FreeSurfer surface coordinates through native scanner RAS into MNI152 RAS space. 
These four transformed meshes serve as the reference surfaces, while the meshes initialized from the predicted segmentation serve as the deformation inputs.

\subsubsection{Segmentation objective}
\label{sec:seg_train}
 
The segmentation network is trained with voxel-wise cross-entropy plus a soft Dice term, \(\mathcal{L}_{\mathrm{seg}} = \mathcal{L}_{\mathrm{CE}} +
\lambda_{\mathrm{Dice}}\mathcal{L}_{\mathrm{Dice}}\). 
The Dice term is averaged over the foreground classes \(c\in\{1,\ldots,8\}\) that are actually present in a given target, so that foreground classes absent from a sample do not contribute to the average; background is excluded throughout. 
Cross-entropy is computed over all nine classes and therefore still penalizes false-positive predictions for absent classes.
 
\subsubsection{Deformation objective}
\label{sec:def_train}
 
The deformation network is trained to refine the initialized meshes toward the aligned reference surfaces while preserving mesh regularity and discouraging anatomically implausible inter-surface contacts. 
All surface-based losses are computed in world coordinates (mm) in MNI152-space.
Throughout, a surface or surface pair is called \emph{valid} when every mesh involved is non-degenerate, that is, has vertices and faces after deformation; invalid entries are skipped so that a single failed mesh does not propagate through the batch. 
Each term below is averaged over the valid predicted--reference surface pairs in the batch.
 
\paragraph{Surface accuracy and mesh regularization}
\label{sec:objective_recon}
 
Surface accuracy is measured by a symmetric Chamfer distance between points sampled from the predicted and reference surfaces, \(\mathcal{L}_{\mathrm{ch}}\). 
The training term uses squared Euclidean distances, which penalize large deviations more strongly; this differs from the Chamfer-\(L_1\) evaluation metric of Section~\ref{sec:metrics}, which uses unsquared nearest-neighbor distances.
 
Mesh geometry is regularized by two terms. 
The edge term \(\mathcal{L}_{\mathrm{edge}}\) is the mean squared difference between the edge lengths of the predicted mesh and those of the corresponding \emph{initialized} mesh under their shared connectivity, so it anchors the deformation to the initialization rather than imposing generic smoothness.
The normal-consistency term \(\mathcal{L}_{\mathrm{normal}}\) penalizes discrepancy between the unit normals of adjacent faces of the predicted mesh. For each pair of faces sharing an edge, it penalizes \(1-\mathbf{n}_{f}^{\top}\mathbf{n}_{f'}\), where the unit face normals are oriented consistently across that edge. 
The penalties are averaged over adjacent-face pairs within each mesh and then over the predicted surfaces in the batch. This discourages abrupt changes in face orientation.

\paragraph{Inter-surface separation penalty}
\label{sec:objective_hd}
 
This objective discourages anatomically relevant surface pairs from approaching one another. 
For two predicted meshes \(\hat{S}_{a}\) and \(\hat{S}_{b}\), let \(d(x,\hat{S})\) be the distance from a point \(x\) to the nearest face of \(\hat{S}\).
Distances are measured in both directions, from the samples of \(\hat{S}_{a}\) to \(\hat{S}_{b}\) and from the samples of \(\hat{S}_{b}\) to \(\hat{S}_{a}\). The two sets are pooled and a single low quantile \(Q_{p}\) is taken over the pooled set, giving a robust estimate of how closely the two surfaces approach.
Falling short of a margin \(m>0\) is then penalized linearly:
\begin{equation}
\begin{split}
\mathrm{Sep}_{p}(\hat{S}_{a},\hat{S}_{b})
&= Q_{p}\Bigl(\{d(x,\hat{S}_{b}):x\in P(\hat{S}_{a})\}\\
&\qquad\ \cup\ \{d(y,\hat{S}_{a}):y\in P(\hat{S}_{b})\}\Bigr),\\
\mathcal{L}_{\mathrm{sep}}^{(p,m)}(\hat{S}_{a},\hat{S}_{b})
&= \mathrm{ReLU}\!\left(m-\mathrm{Sep}_{p}(\hat{S}_{a},\hat{S}_{b})\right).
\end{split}
\label{eq:separation}
\end{equation}
A low quantile rather than the minimum is used because the minimum assigns gradient to a single vertex pair and is sensitive to one poorly shaped triangle, whereas a quantile distributes gradient over the closest few percent of samples.
 
The penalty is applied with its own quantile and margin to the two ipsilateral white--pial pairs, giving \(\mathcal{L}_{\mathrm{wp}}\), and to the left--right pial pair, giving \(\mathcal{L}_{\mathrm{lr}}\). 
No corresponding left--right WM term is used, because the midline gap created during initialization (Section~\ref{sec:surface_initialization}) already separates the two WM surfaces. 
Quantiles and margins were selected empirically on the validation set (Table~\ref{tab:hyperparams}).

\paragraph{White--pial nesting penalty}
\label{sec:objective_signed}
 
Separation alone is not sufficient, for two reasons. 
It is unsigned, so it is satisfied by any well-spaced configuration, including one in which a WM surface lies outside its pial surface. 
It is also bounded: because \(\mathrm{Sep}_{p}\ge 0\), the penalty in Eq.~\eqref{eq:separation} can never exceed \(m\), so two surfaces that merely touch and two that interpenetrate deeply incur the same loss and the same gradient magnitude. 
Separation therefore discourages surfaces from approaching, but carries no information about the severity of a crossing once one has occurred.
 
The second penalty supplies the missing sign and is unbounded in the depth of a violation. 
For a sampled point \(x\) on one surface, let \(\pi(x)\) be its nearest \emph{vertex} on the opposite mesh and \(n(\pi(x))\) the corresponding vertex normal; the signed separation is \(s(x)=\langle x-\pi(x),\,n(\pi(x))\rangle\). 
Under the outward-normal convention inherited from the initialized meshes, each sampled point should lie on its expected side of the opposite surface by at least a margin \(m_{\mathrm{nest}}>0\): WM samples inside the pial surface, so that \(s\le-m_{\mathrm{nest}}\), and pial samples outside the WM surface, so that \(s\ge m_{\mathrm{nest}}\). 
Writing \(s_{\mathrm{w}\rightarrow\mathrm{p}}\) and \(s_{\mathrm{p}\rightarrow\mathrm{w}}\) for the two directions, the two hinges penalize the respective shortfalls:
\begin{equation}
\begin{split}
\mathcal{L}_{\mathrm{nest}} = \tfrac{1}{2}\Bigl[
&\mathbb{E}_{x\sim P(\hat{S}_{w})}
\mathrm{ReLU}\!\left(s_{\mathrm{w}\rightarrow\mathrm{p}}(x)+m_{\mathrm{nest}}\right)\\
+\ &\mathbb{E}_{y\sim P(\hat{S}_{p})}
\mathrm{ReLU}\!\left(m_{\mathrm{nest}}-s_{\mathrm{p}\rightarrow\mathrm{w}}(y)\right)
\Bigr],
\end{split}
\label{eq:nesting}
\end{equation}
averaged over the valid ipsilateral pairs \((w,p)\). 
Unlike Eq.~\eqref{eq:separation}, which reduces each pair to one quantile, this term is a mean over all sampled points, so every violating point contributes gradient.
 

\paragraph{Final objective}
\label{sec:objective_total}
 
The deformation objective is the weighted sum
\begin{equation}
\begin{split}
\mathcal{L}_{\mathrm{def}} ={}&
\lambda_{\mathrm{ch}}\mathcal{L}_{\mathrm{ch}}
+ \lambda_{\mathrm{edge}}\mathcal{L}_{\mathrm{edge}}
+ \lambda_{\mathrm{normal}}\mathcal{L}_{\mathrm{normal}}\\
&+ \lambda_{\mathrm{wp}}\mathcal{L}_{\mathrm{wp}}
+ \lambda_{\mathrm{lr}}\mathcal{L}_{\mathrm{lr}}
+ \lambda_{\mathrm{nest}}\mathcal{L}_{\mathrm{nest}}.
\end{split}
\label{eq:total_loss}
\end{equation}
All weights except \(\lambda_{\mathrm{ch}}\) are ramped linearly from zero to their target values over the first \(T_{\mathrm{w}}\) epochs, so that the Chamfer term establishes geometric alignment before the regularization and separation terms reach full strength. 
Values are given in Table~\ref{tab:hyperparams}.

\section{Experiments}
 
\subsection{Datasets}
\label{sec:datasets}
 
SimCortex v2 was trained using the Human Connectome Project Young Adult (HCP-YA)~\citep{van2013wu} dataset and the Open Access Series of Imaging Studies (OASIS-1)~\citep{marcus2007open}, and was further evaluated on the UCLA Consortium for Neuropsychiatric Phenomics (CNP; OpenNeuro ds000030)~\citep{gorgolewski2017preprocessed} dataset and 12 additional independent datasets from OpenNeuro~\citep{ds000115:00001, ds000144:00002, ds001486:1.3.1, ds001748:1.0.4, ds002862:1.0.1, ds002424:1.2.0, ds002886:1.1.0, ds003499:1.0.1, ds003568:1.0.4, ds003763:1.0.4, ds005234:2.2.0, ds006067:2.0.0}. For all datasets, reference segmentations and cortical surfaces were generated with the FreeSurfer 7.4.1 cross-sectional pipeline.
 
\paragraph{Training and validation data}
HCP-YA is a cohort of healthy young adults aged 22--35 years. From a subset of 100 unrelated subjects, one T1w scan per subject was used. Seventy subjects were assigned to training, 15 to validation, and the remaining 15 were held out for evaluation. OASIS-1 is a cross-sectional dataset spanning adulthood and aging, including cognitively normal individuals and participants with Alzheimer's disease. After excluding one subject with incomplete processing, 415 subjects aged 18--96 years were available. One T1w scan per subject was selected; 291 subjects were used for training, 62 for validation, and the remaining 62 were held out for testing. Splits were created randomly at the subject level within each dataset.
 
\paragraph{Test and generalization data}
The evaluation framework included both in distribution and external test sets. For the in-distribution test set, 40 cases were randomly selected from the pooled held-out HCP-YA and OASIS-1 subjects, comprising 15 HCP-YA and 25 OASIS-1 participants and spanning ages 20--89 years. To assess external generalization, we further evaluated the model on 40 held-out cases from CNP and from each of the 12 additional OpenNeuro datasets. The benchmark therefore comprises 14 cohorts and 560 cases in total: the pooled in-distribution set, and 13 external cohorts contributing 520 cases that were never seen during training.
Candidate cases were required to have a valid T1w image and complete FreeSurfer reference surfaces. These external cohorts were selected to provide a diverse benchmark spanning different age ranges, clinical populations, and scanner vendors, as summarized in Table~\ref{tab:external_datasets}. All test cohorts were strictly held out from model training, validation, and hyperparameter selection.
 
\begin{table*}[t]
\centering
\caption{External datasets used exclusively for testing. Forty eligible
subjects were randomly selected from each dataset for evaluation in this study.}
\label{tab:external_datasets}
 
  \renewcommand{\arraystretch}{1.08}
  \setlength{\tabcolsep}{4pt}
 
  \begingroup
  \resulttablefont
 
\begin{tabular}{l >{\centering\arraybackslash}p{1.5cm} >{\centering\arraybackslash}p{1.7cm} >{\centering\arraybackslash}p{1.8cm} >{\raggedright\arraybackslash}p{4.5cm} >{\raggedright\arraybackslash}p{4.5cm}}
\toprule
Dataset & \ Total subjects & Age range & Subjects used & Population & Scanner \\
\midrule
CNP (ds000030) & 272 & 21--50 & 40 & Healthy / ADHD / Bipolar / Schizophrenia & Siemens Magnetom TrioTim 3T \\
ds000115 & 99  & 11--30 & 40 & Healthy / Schizophrenia & Siemens Magnetom TrioTim 3T\\
ds000144 & 45  & 6--10  & 40 & Anxious children & GE Discovery MR750 / Signa Excite \\
ds001486 & 195 & 8--15  & 40 & Healthy & Siemens Magnetom TrioTim 3T\\
ds001748 & 62  & 10--35 & 40 & Healthy & Siemens Magnetom TrioTim / Prisma 3T\\
ds002424 & 79  & 8--12  & 40 & Healthy / ADHD & Siemens Magnetom TrioTim 3T\\
ds002862 & 71  & 8--10  & 40 & Dyslexia & Siemens Magnetom Skyra 3T\\
ds002886 & 56  & 8--15  & 40 & Healthy & Siemens Magnetom TrioTim 3T\\
ds003499 & 93  & 8--25  & 40 & Healthy & Siemens Magnetom Prisma 3T\\
ds003568 & 49  & 12--19 & 40 & Healthy / Depression & GE Discovery MR750 3T\\
ds003763 & 62  & 18--65 & 40 & Healthy adults without cardiovascular events & Siemens Magnetom Verio 3T \\
ds005234 & 74  & 6--13  & 40 & Healthy / Autism & Siemens Magnetom Verio 3T\\
ds006067 & 118 & 18--40 & 40 & Healthy adults performing a think-aloud task & Philips Ingenia Elition 3T \\
\bottomrule
\end{tabular}
 
  \endgroup
\end{table*}
 
\subsection{Implementation details}
\label{sec:implementation}
\begin{table*}[t]
\centering
\small
\caption{Preprocessing, surface initialization, and training hyperparameters.
Loss-weight superscripts ($0$) denote target weights reached after a 25-epoch
linear warm-up of all non-Chamfer terms.}
\label{tab:hyperparams}
\begin{tabular}{@{}ll@{\hspace{2em}}|ll@{}}
\toprule
Parameter & Value & Parameter & Value \\
\midrule
\multicolumn{2}{@{}l|}{\emph{Preprocessing}} & \multicolumn{2}{l}{\emph{Deformation network}} \\
N4 shrink / spline        & 4 / 200                       & Input                         & $184\times224\times184$, 2 ch. \\
Registration              & affine to MNI152 1\,mm        & MRI encoder widths            & $[8,16,32,64,128,128]$ \\
Reference grid            & $182\times218\times182$       & Ribbon ratio $\rho$ / depth   & 0.5 / 6 \\
\cmidrule(r){1-2}
\multicolumn{2}{@{}l|}{\emph{Surface initialization}}      & Norm / activation             & GroupNorm / LeakyReLU(0.2) \\
Dilation gap $g$          & 1 voxel                       & Dropout                       & 0.1 (3 enc., 2 dec.) \\
SDF smoothing $\sigma_{\mathrm{sdf}}$ & 0.5               & Fusion-gate init              & $-3$ \\
Topology thr.\ $T_{\mathrm{topo}}$ & 16.0                 & SVF $\sigma$ / integ.\ $N$    & 1 / 8 \\
Laplacian iters $n_{\mathrm{smooth}}$ & 1                 & Optimizer                     & AdamW, lr $10^{-4}$, wd $10^{-4}$ \\
WM search $\tau_0,\Delta,\tau_{\min}$ & $-0.2,-0.08,-3.0$ & Batch / accum.\ / clip        & 1 / 4 / 1.0 \\
WM inset $\delta_{\mathrm{wm}}$ & 1.0 voxel               & Points (train / val)          & 180{,}000 / 150{,}000 \\
Pial search $\tau^{\min},\tau^{\max},\Delta$ & $1.8,2.7,0.1$ & Epochs / val         & $\le$600 / every 5 \\
Pial floor                & 0.1 voxel                     & LR sched.                     & Plateau $\times0.5$, pat.\ 4 \\
Ribbon sigmoid $\beta$    & 1.0                           & Early stop                    & pat.\ 15, $\delta=3\!\times\!10^{-4}$ \\
\cmidrule(r){1-2}\cmidrule(l){3-4}
\multicolumn{2}{@{}l|}{\emph{Segmentation network}}        & \multicolumn{2}{l}{\emph{Deformation loss weights}} \\
Feature widths            & $[16,32,64,128,128]$          & $\lambda_{\mathrm{ch}},\lambda^0_{\mathrm{edge}},\lambda^0_{\mathrm{normal}}$ & $1.0, 2.5, 0.25$ \\
Norm / dropout            & instance / 0.10               & $\lambda^0_{\mathrm{wp}},\lambda^0_{\mathrm{nest}},\lambda^0_{\mathrm{lr}}$   & $5.5, 8.0, 1.5$ \\
Optimizer                 & AdamW, lr $10^{-4}$, wd $10^{-5}$ & WP: $p,m$, pts            & 0.05, 0.75\,mm, 50k \\
Batch / schedule          & 2 / cosine to $10^{-6}$       & Nest: $m$, pts                & 0.55\,mm, 50k \\
Epochs / val              & 1000 / every 5                & LR: $p,m$, pts                & 0.05, 1.0\,mm, 30k \\
Loss                      & Dice+CE, $\lambda_{\mathrm{Dice}}=1.0$ & & \\
\bottomrule
\end{tabular}
\end{table*}
All models were implemented in PyTorch, with surface-based losses and mesh sampling operations implemented using PyTorch3D~\citep{ravi2020pytorch3d}.
All hyperparameters for preprocessing, surface initialization and training are given in Table~\ref{tab:hyperparams}; this section describes the procedures and choices that the table does not capture. The complete configuration files and run scripts used for the reported experiments are released with the code repository.
 
\paragraph{Reference segmentations and surfaces}
Reference anatomical labels and cortical surfaces for every cohort were produced with the FreeSurfer 7.4.1 cross-sectional pipeline \citep{fischl2012freesurfer}, run with \texttt{recon-all -all} using default parameters, on a single T1w volume per subject. 
The same FreeSurfer version was used for all 14 cohorts, so that reference geometry is consistent across the benchmark. 
No manual editing or topological correction was applied to the FreeSurfer output: the references are therefore an unedited silver standard rather than a manually verified ground truth.
Subjects whose reconstruction did not complete were excluded.
 
From each reconstruction we used the \texttt{aparc+aseg} volume, grouped into the nine anatomical classes of Table~\ref{tab:seg9_mapping} to form the segmentation targets. 
The \texttt{lh.white}, \texttt{rh.white}, \texttt{lh.pial}, and \texttt{rh.pial} surfaces were used as the reference surfaces for deformation-network training.

\begin{table*}[t]
  \centering
  \caption{Mapping from FreeSurfer \texttt{aparc+aseg} labels to the nine
  segmentation classes used by SimCortex v2. Voxels whose label is not listed
  are assigned to background. FreeSurfer WM hypointensity labels
  77 and 80 carry no hemisphere and are split into the left (class~1) or
  right (class~2) inner-cerebrum class using \texttt{filled.mgz}
  (255~$\rightarrow$~left, 127~$\rightarrow$~right).}
  \label{tab:seg9_mapping}

  \renewcommand{\arraystretch}{1.08}
  \setlength{\tabcolsep}{4pt}

  \begingroup
  \resulttablefont

  \begin{tabular*}{\textwidth}{
    @{\extracolsep{\fill}}
    c
    p{0.30\textwidth}
    p{0.57\textwidth}
    @{}
  }
    \toprule

    Class
    & Region
    & FreeSurfer \texttt{aparc+aseg} labels \\

    \midrule

    0
    & Background / non-target tissue
    & all labels not listed below \\

    1
    & Left inner cerebrum / WM
    & 2, 5, 10, 11, 12, 13, 26, 28, 30, 31
      (+77/80 if left) \\

    2
    & Right inner cerebrum / WM
    & 41, 44, 49, 50, 51, 52, 58, 60, 62, 63
      (+77/80 if right) \\

    3
    & Left cortical ribbon
    & 3, 1000--1003, 1005--1035 \\

    4
    & Right cortical ribbon
    & 42, 2000--2003, 2005--2035 \\

    5
    & Left hippocampus--amygdala
    & 17, 18 \\

    6
    & Right hippocampus--amygdala
    & 53, 54 \\

    7
    & Left lateral ventricle
    & 4 \\

    8
    & Right lateral ventricle
    & 43 \\

    \bottomrule
  \end{tabular*}

  \endgroup
\end{table*}
 
\paragraph{Preprocessing}
N4 bias-field correction was applied to the native T1w image using an automatically estimated brain mask. The corrected image was then affinely registered to the MNI152 T1 1\,mm reference using ANTsPy, and all subsequent processing was carried out on that reference grid at 1\,mm isotropic resolution.
 
\paragraph{Segmentation network}
The segmentation network (Section~\ref{sec:segmentation}) was trained on the HCP-YA and OASIS-1 training splits with mixed-precision training and gradient clipping at 1.0.
 
On-the-fly spatial augmentation used random affine transformations with probability 0.5, rotations of up to approximately \(\pm15^\circ\) about each axis, and scaling of up to \(\pm10\%\); translations were disabled because the images were already aligned to the common MNI152 frame. 
Intensity augmentation included random contrast adjustment with probability 0.3 and gamma range \([0.7,1.5]\), smooth bias-field perturbation with probability 0.3, and additive Gaussian noise with probability 0.1 and standard deviation 0.05.
 
The checkpoint with the highest validation mean foreground Dice score was used for inference; it was obtained at epoch 925 with a validation Dice of 0.9261.
 
\paragraph{Deformation network}
Convolutional blocks of the deformation network (Section~\ref{sec:deformation}) used Group Normalization with up to eight groups. 
Weight decay was applied to convolutional and projection weights only; biases, normalization parameters, other one-dimensional parameters and the ribbon fusion-gate logits were excluded from it.
 
Training used gradient accumulation over four iterations and gradient clipping at 1.0. 
A \texttt{ReduceLROnPlateau} scheduler monitored the validation Chamfer RMSE, with cooldown 1, and early stopping used the same criterion. 
Training terminated at epoch 540, and the minimum-validation-RMSE checkpoint was obtained at epoch 465 with a validation Chamfer RMSE of \(1.1997\,\mathrm{mm}\). 
That checkpoint was used for all reported inference and evaluation results.
 
Spatial augmentation was applied consistently to the input volume, initialized surfaces, and reference surfaces using a shared random 3D affine transform with probability 0.3, rotations up to \(\pm3^\circ\), isotropic
scaling up to \(\pm2\%\), and translations up to \(\pm1\,\mathrm{mm}\).
Image-only augmentation of the MRI channel was applied with probability 0.3 and included smooth multiplicative bias-field perturbation with strength 0.2, global gain and brightness changes of up to \(\pm0.1\), and additive Gaussian noise with standard deviation 0.05.
 
No smoothing, remeshing, topology repair, or self-intersection removal was applied to the predicted surfaces after deformation. 
 
\paragraph{Baselines}
We evaluated SimCortex v2 against five learning-based cortical surface reconstruction methods: CF \citep{lebrat2021corticalflow}, CFPP \citep{santa2022corticalflow++}, CortexODE \citep{ma2022cortexode}, V2C
\citep{bongratz2022vox2cortex}, and V2C-Flow \citep{bongratz2024neural}, as well as the FreeSurfer~7.4.1
cross-sectional pipeline \citep{fischl2012freesurfer}. 
The learning-based baselines were run using their released pretrained weights and their native preprocessing and initialization procedures, without retraining or fine-tuning on our evaluation data. 
This evaluates each method under its intended released inference pipeline rather than forcing a common preprocessing strategy, although we recognize some of the baselines may be at a disadvantage for some accuracy metrics as they may have been trained with different versions of FreeSurfer silver standard references.

All methods were scored on the same 560-case benchmark against the same FreeSurfer references using the same evaluation implementation and metric definitions.
 
We also use FreeSurfer as a baseline method, although only for geometric validity because its surfaces serve as the reference for all accuracy metrics. 
FreeSurfer is therefore excluded from the surface-distance and thickness comparisons, for which it would be trivially exact, and appears only in the self-intersection and collision results.

\subsection{Evaluation metrics}
\label{sec:metrics}
 
We evaluated SimCortex v2 in terms of geometric accuracy, surface self-intersection, cortical thickness agreement, and inter-surface collision statistics. 
Before evaluation, predicted and reference surfaces were represented in each subject's scanner RAS coordinate system. 
All geometric distances were therefore computed in physical world coordinates and reported in millimeters.
For the surface-distance metrics, 150{,}000 points were sampled uniformly from each surface, while 50{,}000 WM points per hemisphere were used for cortical thickness estimation.
 
\paragraph{Location accuracy metrics}
For each predicted and reference surface (left/right WM and pial), we computed bidirectional distance-based metrics. 
Our primary surface-distance metric is the Average Symmetric Surface Distance (ASSD), defined as the average of the mean point-to-surface (point-to-triangle) distances in the two directions. 
We also report the 90th-percentile Hausdorff distance (HD90), defined as the larger of the two directional 90th-percentile point-to-surface distances. 
Compared with the classical maximum Hausdorff distance, HD90 is less sensitive to isolated outliers. 
For continuity with prior point-cloud-based evaluations, we additionally report a symmetric Chamfer-\(L_1\) distance, computed from nearest-neighbor distances between the two sampled point clouds and averaged
over both directions. 
Unlike ASSD and HD90, which use point-to-triangle distances, Chamfer-\(L_1\) is therefore a point-to-point measure.

Cortical thickness agreement is evaluated separately for each hemisphere by sampling points on the WM surface and computing their point-to-surface distances to the corresponding pial mesh. 
The mean of these distances defines the predicted mean white-to-pial thickness. 
The same procedure is applied to the reference surfaces, from which we compute the per-hemisphere absolute thickness error and signed thickness bias in millimeters. 
This metric represents mean nearest-surface white-to-pial separation rather than a vertex-wise anatomical correspondence measure.

\paragraph{Geometric validity metrics}
To quantify self-intersection, we measure the self-intersection fraction (SIF) on each predicted surface using PyMeshLab~\citep{pymeshlab}. SIF is reported as the percentage of faces flagged as self-intersecting by PyMeshLab's per-face self-intersection selection filter relative to the total number of faces on that surface.
 
Inter-surface collisions are assessed using triangle-level collision queries issued directly to the Flexible Collision Library (FCL) \citep{pan2012fcl} through its Python bindings. 
The same queries are used by the collision-aware isolevel selection of Section~\ref{sec:surface_initialization}, so initialization and evaluation apply an identical test. 
For each subject, we test all six pairwise combinations of the four predicted cortical surfaces: left--right pial, left--right WM, left WM--left pial, right WM--right pial, left WM--right pial, and right WM--left pial. 
Contact queries are repeated with larger contact budgets when necessary to avoid truncating the set of intersecting faces.
 
From these pairwise tests, we derive a per-surface \emph{union} collision measure. 
For each surface \(s\) with face set \(F_s\), let \(C_s\) denote the set of faces that intersect any of the other three cortical surfaces,
\[
C_s
=
\left\{
f\in F_s :
f \text{ intersects any surface } s' \neq s
\right\}.
\]
Each face is counted only once, regardless of how many other surfaces it intersects. The corresponding collision percentage is 
\[
c_s
=
100\times\frac{|C_s|}{|F_s|}.
\]
The case-level union collision measure is the mean over the four cortical surfaces,
\[
\mathrm{Collision}
=
\frac{1}{4}\sum_s c_s.
\]

\section{Results}
\label{sec:results}

\subsection{Overall comparison with state-of-the-art methods}
\label{sec:results_main}

We compared SimCortex v2 against five recent learning-based cortical surface reconstruction methods: CF, CFPP, CortexODE, V2C, and V2C-Flow. 
We additionally report the self-intersection and inter-surface collision statistics of the
FreeSurfer reference surfaces. 
Performance was evaluated across three complementary categories of metrics: surface-distance accuracy (ASSD, HD90,
and Chamfer-$L_1$), cortical thickness absolute error, and geometric validity (SIF and union collision percentage). Since the FreeSurfer surfaces served as the silver-standard reference for the surface-distance and thickness metrics, FreeSurfer was excluded from those two comparisons.

As shown in Table~\ref{tab:overall_results}, SimCortex v2 achieved the lowest mean ASSD, HD90, and Chamfer-$L_1$ values and the lowest union collision percentage among all evaluated methods. 
SimCortex v2 was completely collision-free in 92.14\% (516/560) of the evaluation cases, whereas every other learning-based method exhibited nonzero collision in every case. 
For cortical thickness absolute error, V2C-Flow ranked first, with SimCortex v2 ranking second. The FreeSurfer reference surfaces had the lowest overall SIF, while SimCortex v2 achieved the lowest SIF among the
learning-based methods. 
An ordinal summary across all metrics is provided in Table~S2 of the Supplementary Material.

Figure~\ref{fig:assd_splitviolin} provides a complementary view of ASSD by comparing the ASSD distribution of SimCortex v2 against each competing learning-based method. SimCortex v2 achieved significantly lower
ASSD than CF, CFPP, V2C, and CortexODE (two-sided Wilcoxon signed-rank, all $p<0.001$). 
Relative to V2C-Flow, SimCortex v2 had a slightly lower mean
ASSD ($0.253$ vs.\ $0.255$\,mm), whereas the paired comparison favored
V2C-Flow, with a median paired difference of approximately $0.016$\,mm in
its favor ($p<0.001$). SimCortex v2 nevertheless showed lower variability
(SD $0.067$ vs.\ $0.095$\,mm) and substantially lower SIF and union
collision values.

Among the remaining learning-based methods, CFPP improved over CF in ASSD,
HD90, Chamfer-$L_1$, cortical thickness error, and SIF, although its union
collision percentage was higher. V2C had larger surface-distance errors than
CF and CFPP and also showed a comparatively high SIF. CortexODE had the
largest mean surface-distance errors and the largest standard deviations for
all three surface-distance metrics, indicating greater variability across the
560 evaluation cases. The FreeSurfer reference surfaces had the
lowest SIF but the highest union collision percentage overall. This contrast
shows that SIF, which measures self-intersections within individual surfaces,
and inter-surface collision, which measures intersections between distinct
cortical surfaces, characterize different forms of geometric inconsistency
and should therefore be evaluated separately.

\begin{table*}[t]
  \centering
    \caption{Overall cortical surface reconstruction performance across 560 evaluation cases. Continuous metrics are reported as mean $\pm$ SD and lower values are better. Coll (\%) is the percentage of cases with at least one union collision. Union collision is the mean, over the four cortical surfaces, of the percentage of each surface's faces that intersect any of the other three surfaces, with intersections evaluated across all six surface pairs. Because the FreeSurfer surfaces serve as the silver-standard reference for surface-distance and thickness metrics, FreeSurfer is not included in those comparisons; its SIF and collision values describe the intrinsic geometric validity of the reference surfaces themselves. Best results are in bold and second-best results are underlined.}
  \label{tab:overall_results}
  \renewcommand{\arraystretch}{1.08}
  \setlength{\tabcolsep}{4pt}

  \begingroup
  \resulttablefont

\begin{tabular}{l c c c c c c c}
    \toprule
    Method
    & ASSD (mm)
    & HD90 (mm)
    & Chamfer (mm)
    & Thickness abs.\ err.\ (mm)
    & SIF (\%)
    & Coll union (\%)
    & Coll (\%) \\
    \midrule
    V2C\text{-}Flow
    & $\underline{0.255 \pm 0.095}$
    & $\underline{0.600 \pm 0.290}$
    & $\underline{0.534 \pm 0.082}$
    & $\mathbf{0.105 \pm 0.065}$
    & $2.843 \pm 0.890$
    & $1.423 \pm 0.171$
    & $100.00$ \\
    CFPP
    & $0.358 \pm 0.189$
    & $0.837 \pm 0.612$
    & $0.621 \pm 0.173$
    & $0.219 \pm 0.158$
    & $0.343 \pm 0.265$
    & $0.944 \pm 0.208$
    & $100.00$ \\
    CF
    & $0.435 \pm 0.144$
    & $1.070 \pm 0.558$
    & $0.670 \pm 0.132$
    & $0.271 \pm 0.156$
    & $0.770 \pm 0.298$
    & $\underline{0.762 \pm 0.143}$
    & $100.00$ \\
    V2C
    & $0.482 \pm 0.155$
    & $1.245 \pm 0.516$
    & $0.699 \pm 0.142$
    & $0.400 \pm 0.170$
    & $1.116 \pm 0.533$
    & $0.910 \pm 0.091$
    & $100.00$ \\
    CortexODE
    & $1.124 \pm 1.570$
    & $3.663 \pm 6.052$
    & $1.346 \pm 1.548$
    & $0.296 \pm 0.255$
    & $0.217 \pm 0.138$
    & $1.403 \pm 0.796$
    & $100.00$ \\
    FreeSurfer
    & --
    & --
    & --
    & --
    & $\mathbf{0.003 \pm 0.004}$
    & $6.667 \pm 0.584$
    & $100.00$ \\
    \midrule
    SimCortex v2
    & $\mathbf{0.253 \pm 0.067}$
    & $\mathbf{0.586 \pm 0.228}$
    & $\mathbf{0.529 \pm 0.060}$
    & $\underline{0.124 \pm 0.065}$
    & $\underline{0.044 \pm 0.014}$
    & $\mathbf{9.04\times10^{-5} \pm 6.47\times10^{-4}}$
    & $\mathbf{7.86}$ \\
    \bottomrule
  \end{tabular}%
\endgroup
\end{table*}

\begin{figure*}[t]
  \centering
  \includegraphics[width=\textwidth]{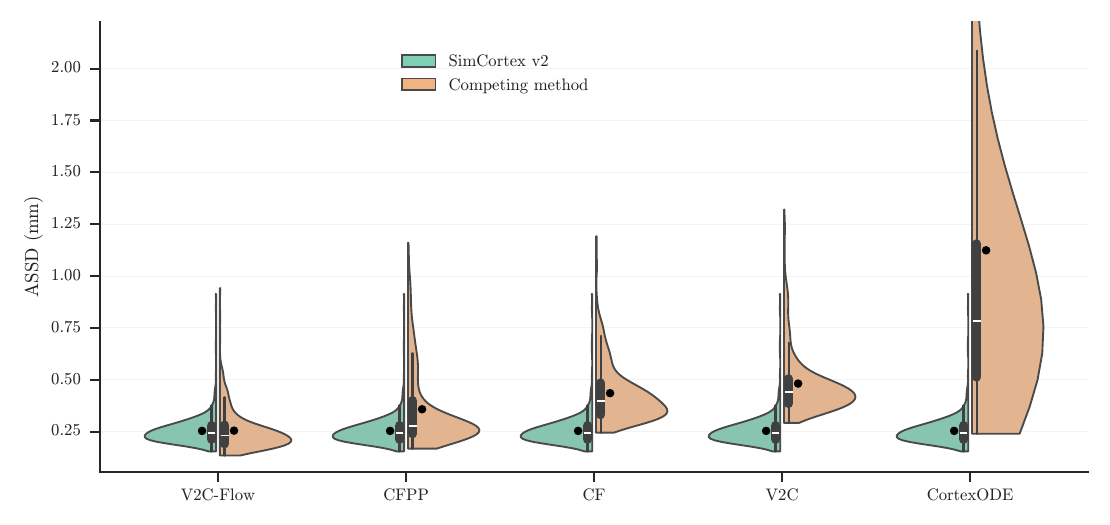}
    \caption{Distributions of ASSD for SimCortex v2 against each competing
    learning-based method. Each column shows a paired comparison over the same
    560 evaluation cases: the left half of each violin represents SimCortex v2
    and the right half represents the competing method indicated on the
    horizontal axis. Inner boxes show the median and interquartile range, and
    black dots indicate the means. Columns are ordered by the competing method's
    median ASSD. For visual readability, the displayed range is clipped to the
    0.5th--99.5th percentiles of the pooled plotted values; clipping affects
    visualization only, and all summary statistics shown in the figure are
    computed from the full, unclipped data. Lower ASSD values are better.}
  \label{fig:assd_splitviolin}
\end{figure*}

\subsubsection{Computational efficiency}
\label{sec:results_runtime}

Runtime results are summarized in Table~\ref{tab:runtime_comparison}.
The two measurements capture different computational aspects of each method.
When only the surface generation model computation was measured, SimCortex v2 had the lowest mean inference time at \(0.594\)~s for all four surfaces, followed by V2C at \(0.685\)~s and CortexODE at \(0.784\)~s.
The corresponding times were \(3.491\)~s for CFPP, \(5.386\)~s for CF, and \(6.792\)~s for V2C-Flow.

The end-to-end comparison showed a different pattern because it additionally includes each method's required preprocessing, registration, initialization, and reconstruction stages. 
CortexODE had the lowest end-to-end runtime at \(24.150\)~s, followed by V2C (\(133.994\)~s), V2C-Flow (\(138.259\)~s),
CFPP (\(140.536\)~s), CF (\(162.448\)~s), and SimCortex v2 (\(174.248\)~s). 
Thus, although the learned surface deformation stage of SimCortex v2 is fast, its current end-to-end runtime is dominated by the preceding preprocessing and topology-aware surface initialization stages.

\begin{table}[t]
  \centering
  \caption{Runtime comparison across 28 cases comprising two subjects
    from each of the 14 evaluation datasets. Values are mean $\pm$ SD in seconds
    and correspond to generation of all four cortical surfaces. Surface-generation
    inference measures the GPU-synchronized model computation required to generate
    all four cortical surfaces from model-ready inputs. End-to-end reconstruction measures total time through each method's required preprocessing and reconstruction pipeline
    to the four final surfaces. Benchmarks were run on one NVIDIA RTX A6000 GPU (48 GB) on a workstation with an Intel Xeon Gold 5218R CPU at 2.10 GHz (20 cores/40 threads) and 128 GiB RAM. Best results are shown in bold and second-best
    results are underlined.}
  \label{tab:runtime_comparison}

  \renewcommand{\arraystretch}{1.10}
  \setlength{\tabcolsep}{5pt}

  \begingroup
  \resulttablefont

  \begin{tabular}{@{}lcc@{}}
    \toprule
    Method
    & \shortstack{Surface generation\\inference (s)}
    & \shortstack{End-to-end\\reconstruction (s)} \\
    \midrule

    V2C\text{-}Flow
    & $6.792 \pm 0.155$
    & $138.259 \pm 26.927$ \\

    CFPP
    & $3.491 \pm 0.135$
    & $140.536 \pm 29.644$ \\

    CF
    & $5.386 \pm 0.047$
    & $162.448 \pm 22.419$ \\

    V2C
    & $\underline{0.685 \pm 0.016}$
    & $\underline{133.994 \pm 21.164}$ \\

    CortexODE
    & $0.784 \pm 0.102$
    & $\mathbf{24.150 \pm 0.829}$ \\

    \midrule

    SimCortex v2
    & $\mathbf{0.594 \pm 0.005}$
    & $174.248 \pm 18.746$ \\

    \bottomrule
  \end{tabular}

  \endgroup
\end{table}

\subsubsection{Generalization to independent datasets}
\label{sec:results_generalization}

Table~\ref{tab:dataset_assd_collision_means} reports the mean ASSD and
union collision percentage for each method across the 14 evaluation datasets, providing a dataset-level view of surface-distance accuracy and inter-surface collision behavior. Complete per-dataset results for all four metrics, including cortical thickness error and SIF, are given in Table~S1 of the Supplementary Material.

V2C-Flow achieved the lowest mean ASSD on nine datasets, whereas SimCortex v2 achieved the lowest mean ASSD on five and ranked second on the remaining nine. This pattern was also reflected in the pooled domain-level results: across the 520 cases from the 13 external datasets, V2C-Flow had a lower mean ASSD than SimCortex v2 (0.254 vs.\ 0.257\,mm), whereas on the 40 held-out in-distribution HCP-YA/OASIS-1 cases, SimCortex v2 achieved the lower mean ASSD (0.207 vs.\ 0.268\,mm). The two methods therefore differ in how they transfer: SimCortex v2 is more accurate in distribution and loses \(0.050\,\mathrm{mm}\) externally, whereas V2C-Flow performs slightly better externally than in distribution. Part of the in-distribution advantage is thus attributable to the match between the training cohorts and the reference surfaces used here, and the in-distribution estimate rests on only 40 cases.

SimCortex v2 obtained the lowest union collision percentage on all 14 datasets. Its dataset-level means ranged from \(0\) to \(4.15\times10^{-4}\%\), whereas the lowest collision percentages among the other learning-based methods ranged from \(0.615\%\) to \(0.885\%\). The same pattern held after pooling by domain: SimCortex v2 achieved \(9.45\times10^{-5}\%\) union collision across the 520 external cases and \(3.72\times10^{-5}\%\) across the 40 held-out in-distribution cases. Thus, the near-zero overall collision percentage reported in Table~\ref{tab:overall_results} was consistently observed across datasets rather than being driven by a small subset of cases.

\begin{table*}[t]
  \centering
    \caption{ASSD and union collision percentage for each method per dataset.
    For each case, ASSD is averaged across the four cortical surfaces, and union collision is the mean of the four per-surface union collision percentages. 
    Dataset-level values are averaged over 40 cases. The out of domain (\textbf{OOD overall}) row pools the 520 cases from the 13 external datasets, whereas the in distribution \textbf{ID overall} row contains the 40 held-out HCP-YA/OASIS-1 cases in \texttt{hcp\_oasis}. FreeSurfer provides the reference surfaces for ASSD and is therefore excluded from the ASSD comparison. Lower values are better for both metrics. Best and second-best values are shown in bold and underlined, respectively. Very small nonzero collision percentages are reported in scientific notation.}
  
  \label{tab:dataset_assd_collision_means}
  \renewcommand{\arraystretch}{1.08}
  \setlength{\tabcolsep}{4pt}
  \begingroup
  \resulttablefont
  \begin{tabular}{l cc cc cc cc cc cc cc}
    \toprule
    & \multicolumn{2}{c}{V2C\text{-}Flow}
    & \multicolumn{2}{c}{CFPP}
    & \multicolumn{2}{c}{CF}
    & \multicolumn{2}{c}{V2C}
    & \multicolumn{2}{c}{CortexODE}
    & \multicolumn{2}{c}{FreeSurfer}
    & \multicolumn{2}{c}{SimCortex v2} \\
    \cmidrule(lr){2-3}
    \cmidrule(lr){4-5}
    \cmidrule(lr){6-7}
    \cmidrule(lr){8-9}
    \cmidrule(lr){10-11}
    \cmidrule(lr){12-13}
    \cmidrule(lr){14-15}
    Dataset
    & ASSD & Coll.
    & ASSD & Coll.
    & ASSD & Coll.
    & ASSD & Coll.
    & ASSD & Coll.
    & ASSD & Coll.
    & ASSD & Coll. \\

    \midrule
    cnp
    & $\mathbf{0.188}$
    & 1.409
    & 0.244
    & 0.838
    & 0.324
    & $\underline{0.713}$
    & 0.407
    & 0.900
    & 0.504
    & 1.606
    & --
    & 6.770
    & $\underline{0.218}$
    & $\mathbf{0}$ \\
    ds000115
    & $\mathbf{0.185}$
    & 1.340
    & 0.225
    & 0.746
    & 0.311
    & $\underline{0.615}$
    & 0.357
    & 0.901
    & 0.490
    & 1.219
    & --
    & 6.754
    & $\underline{0.206}$
    & $\mathbf{0}$ \\
    ds000144
    & $\mathbf{0.274}$
    & 1.362
    & 0.295
    & 0.962
    & 0.451
    & $\underline{0.855}$
    & 0.513
    & 0.889
    & 0.982
    & 1.301
    & --
    & 6.397
    & $\underline{0.282}$
    & $\mathbf{5.09\times10^{-5}}$ \\
    ds001486
    & $\underline{0.294}$
    & 1.511
    & 0.325
    & 0.816
    & 0.422
    & $\underline{0.671}$
    & 0.469
    & 0.908
    & 0.929
    & 2.028
    & --
    & 6.865
    & $\mathbf{0.287}$
    & $\mathbf{2.98\times10^{-5}}$ \\
    ds001748
    & $\mathbf{0.233}$
    & 1.365
    & 0.253
    & 0.944
    & 0.387
    & $\underline{0.758}$
    & 0.425
    & 0.983
    & 0.653
    & 0.854
    & --
    & 6.548
    & $\underline{0.244}$
    & $\mathbf{9.76\times10^{-5}}$ \\
    ds002424
    & $\underline{0.313}$
    & 1.353
    & 0.424
    & 0.879
    & 0.451
    & $\underline{0.682}$
    & 0.470
    & 0.921
    & 1.071
    & 2.362
    & --
    & 6.805
    & $\mathbf{0.301}$
    & $\mathbf{1.12\times10^{-4}}$ \\
    ds002862
    & $\underline{0.421}$
    & 1.438
    & 0.826
    & 1.417
    & 0.715
    & 1.005
    & 0.480
    & $\underline{0.885}$
    & 1.496
    & 2.053
    & --
    & 6.222
    & $\mathbf{0.284}$
    & $\mathbf{4.58\times10^{-5}}$ \\
    ds002886
    & $\underline{0.267}$
    & 1.561
    & 0.285
    & 0.813
    & 0.395
    & $\underline{0.657}$
    & 0.448
    & 0.925
    & 0.958
    & 2.005
    & --
    & 6.636
    & $\mathbf{0.261}$
    & $\mathbf{6.54\times10^{-5}}$ \\
    ds003499
    & $\mathbf{0.201}$
    & 1.557
    & 0.263
    & 0.852
    & 0.422
    & $\underline{0.779}$
    & 0.406
    & 0.927
    & 0.838
    & 0.859
    & --
    & 6.579
    & $\underline{0.221}$
    & $\mathbf{5.65\times10^{-5}}$ \\
    ds003568
    & $\mathbf{0.220}$
    & 1.407
    & 0.320
    & 0.976
    & 0.406
    & $\underline{0.724}$
    & 0.729
    & 0.859
    & 0.950
    & 1.011
    & --
    & 6.482
    & $\underline{0.252}$
    & $\mathbf{1.18\times10^{-4}}$ \\
    ds003763
    & $\mathbf{0.199}$
    & 1.585
    & 0.295
    & 0.841
    & 0.357
    & $\underline{0.794}$
    & 0.388
    & 0.980
    & 0.883
    & 1.261
    & --
    & 7.267
    & $\underline{0.225}$
    & $\mathbf{1.26\times10^{-4}}$ \\
    ds005234
    & $\mathbf{0.263}$
    & 1.326
    & 0.328
    & 1.054
    & 0.417
    & $\underline{0.855}$
    & 0.478
    & 0.873
    & 0.735
    & 0.961
    & --
    & 6.360
    & $\underline{0.264}$
    & $\mathbf{4.15\times10^{-4}}$ \\
    ds006067
    & $\mathbf{0.239}$
    & 1.330
    & 0.574
    & 1.030
    & 0.568
    & $\underline{0.693}$
    & 0.763
    & 0.871
    & 1.208
    & 1.259
    & --
    & 6.816
    & $\underline{0.295}$
    & $\mathbf{1.11\times10^{-4}}$ \\
    \midrule
    \textbf{OOD overall} ($n=520$)
    & $\mathbf{0.254}$
    & 1.427
    & 0.358
    & 0.936
    & 0.433
    & $\underline{0.754}$
    & 0.487
    & 0.909
    & 0.900
    & 1.444
    & --
    & 6.654
    & $\underline{0.257}$
    & $\mathbf{9.45\times10^{-5}}$ \\

    \textbf{ID overall} ($n=40$)
    & $\underline{0.268}$
    & 1.379
    & 0.353
    & 1.041
    & 0.469
    & 0.872
    & 0.410
    & 0.916
    & 4.041
    & $\underline{0.870}$
    & --
    & 6.831
    & $\mathbf{0.207}$
    & $\mathbf{3.72\times10^{-5}}$ \\
    \bottomrule
  \end{tabular}%
\endgroup
\end{table*}

\subsubsection{Surface location accuracy and geometric validity}
\label{sec:results_surfacewise}

Table~\ref{tab:method_surface_assd_collision} reports ASSD and union collision percentage for each of the four surfaces independently (left/right WM and pial).
For every learning-based method, mean ASSD was higher for the pial surface than for the corresponding WM surface in both hemispheres, indicating greater surface-distance error at the pial boundary.

V2C-Flow achieved the lowest mean ASSD for both WM surfaces
(0.206 and 0.211\,mm for the left and right hemispheres, respectively),
with SimCortex v2 ranking second (0.223 and 0.229\,mm). For the pial
surfaces, this ordering reversed: SimCortex v2 achieved the lowest mean ASSD (0.278 and 0.283\,mm), followed by V2C-Flow (0.296 and 0.305\,mm).

\begin{figure*}[t]
  \centering
  \includegraphics[width=\linewidth]{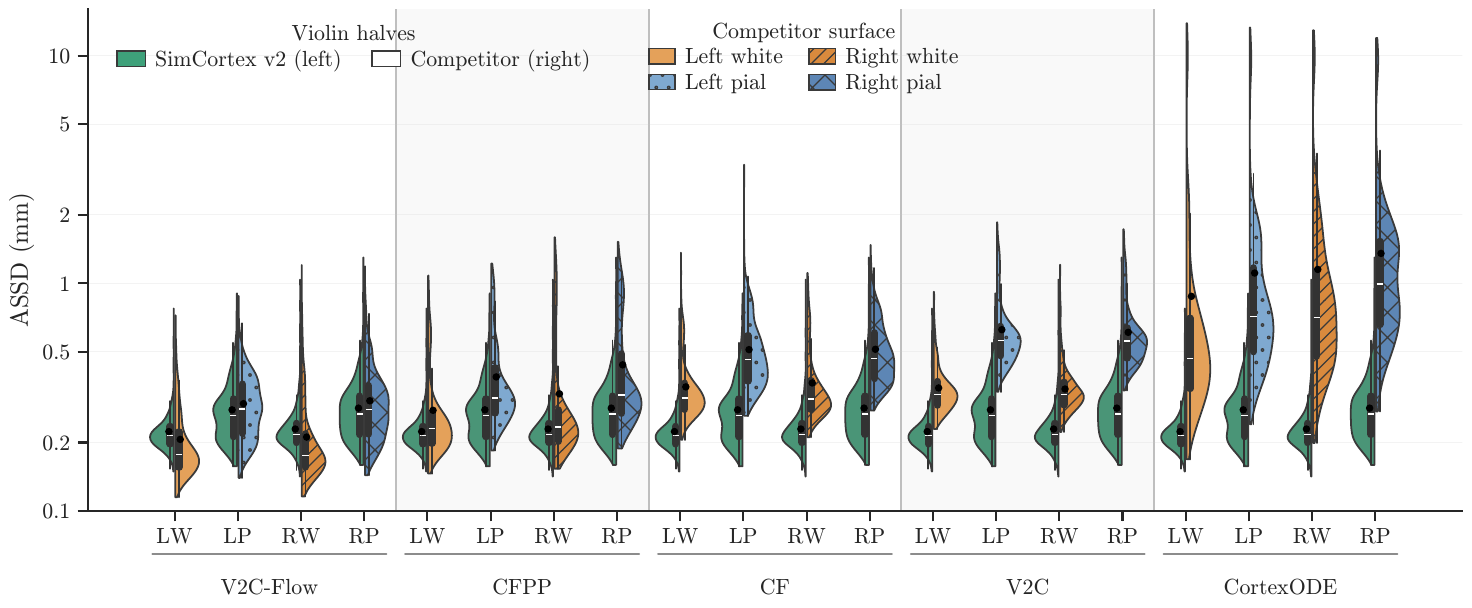}
  \caption{Distributions of ASSD for SimCortex v2 and the five competing learning-based methods. Within each competitor group, the four columns correspond to the
  left WM (LW), left pial (LP), right WM (RW), and right
  pial (RP) surfaces. Each split violin represents a paired comparison over
  the same 560 evaluation cases: the left half (green) shows SimCortex v2 and
  the right half shows the competing method. Inner boxes indicate the median
  and interquartile range, and black dots indicate arithmetic means computed
  in the original metric units. The vertical axis is logarithmic to
  accommodate the larger ASSD range of CortexODE while preserving visibility
  of the distributions of the other methods. Lower ASSD is better.}
  \label{fig:assd_surface}
\end{figure*}

Figure~\ref{fig:assd_surface} provides an illustration of the distribution of these surface-specific differences. 
The distributions show the close agreement between SimCortex v2 and V2C-Flow on the WM surfaces, while
SimCortex v2 shifts toward lower ASSD on both pial surfaces. 
The broader upper tail of CortexODE is also evident across the four surfaces.

SimCortex v2 also achieved the lowest per-surface union collision percentage for all four surfaces. 
Its mean collision percentages ranged from \(1.43\times10^{-5}\%\) to \(1.55\times10^{-4}\%\), more than three orders of magnitude below the next-best value for every surface. 
Thus, SimCortex v2 combined the best pial-surface accuracy with the second-best WM accuracy while maintaining substantially lower inter-surface collision across all four reconstructed surfaces.

\begin{table*}[t]
  \centering
  \caption{ASSD and union collision percentage for each surface type across 560 evaluation cases 
  (mean $\pm$ SD; lower is better). For each surface, the collision
  percentage reflects its union of overlaps with the other three cortical
  surfaces. FreeSurfer, used as the reference for ASSD, is excluded from
  that comparison. L and R denote left and right hemispheres. Best results are in bold; second-best are underlined.}
  \label{tab:method_surface_assd_collision}

  \renewcommand{\arraystretch}{1.08}
  \setlength{\tabcolsep}{4pt}

  \begingroup
  \resulttablefont

  \begin{tabular}{@{}l cc cc@{}}
    \toprule
    & \multicolumn{2}{c}{L WM}
    & \multicolumn{2}{c}{L Pial} \\
    \cmidrule(lr){2-3}
    \cmidrule(lr){4-5}

    Method
    & ASSD (mm) & Coll. (\%)
    & ASSD (mm) & Coll. (\%) \\
    \midrule

    V2C\text{-}Flow
    & $\mathbf{0.206 \pm 0.085}$
    & $1.295 \pm 0.167$
    & $\underline{0.296 \pm 0.097}$
    & $1.580 \pm 0.228$ \\

    CFPP
    & $0.277 \pm 0.147$
    & $0.804 \pm 0.196$
    & $0.388 \pm 0.198$
    & $1.011 \pm 0.226$ \\

    CF
    & $0.351 \pm 0.129$
    & $\underline{0.643 \pm 0.127}$
    & $0.512 \pm 0.215$
    & $\underline{0.753 \pm 0.151}$ \\

    V2C
    & $0.347 \pm 0.093$
    & $0.973 \pm 0.113$
    & $0.626 \pm 0.248$
    & $0.755 \pm 0.118$ \\

    CortexODE
    & $0.878 \pm 1.644$
    & $1.128 \pm 0.639$
    & $1.111 \pm 1.561$
    & $1.310 \pm 0.641$ \\

    FreeSurfer
    & --
    & $6.336 \pm 0.603$
    & --
    & $7.024 \pm 0.605$ \\

    \midrule

    SimCortex v2
    & $\underline{0.223 \pm 0.050}$
    & $\mathbf{1.43\times10^{-5} \pm 1.53\times10^{-4}}$
    & $\mathbf{0.278 \pm 0.090}$
    & $\mathbf{1.04\times10^{-4} \pm 4.59\times10^{-4}}$ \\

    \bottomrule
  \end{tabular}

  \vspace{4pt}

  \begin{tabular}{@{}l cc cc@{}}
    \toprule
    & \multicolumn{2}{c}{R WM}
    & \multicolumn{2}{c}{R Pial} \\
    \cmidrule(lr){2-3}
    \cmidrule(lr){4-5}

    Method
    & ASSD (mm) & Coll. (\%)
    & ASSD (mm) & Coll. (\%) \\
    \midrule

    V2C\text{-}Flow
    & $\mathbf{0.211 \pm 0.105}$
    & $1.276 \pm 0.180$
    & $\underline{0.305 \pm 0.121}$
    & $1.542 \pm 0.243$ \\

    CFPP
    & $0.327 \pm 0.258$
    & $0.875 \pm 0.224$
    & $0.439 \pm 0.267$
    & $1.085 \pm 0.266$ \\

    CF
    & $0.365 \pm 0.152$
    & $\underline{0.735 \pm 0.204}$
    & $0.514 \pm 0.175$
    & $0.918 \pm 0.234$ \\

    V2C
    & $0.343 \pm 0.089$
    & $1.090 \pm 0.109$
    & $0.610 \pm 0.224$
    & $\underline{0.821 \pm 0.127}$ \\

    CortexODE
    & $1.152 \pm 1.648$
    & $1.480 \pm 1.034$
    & $1.355 \pm 1.541$
    & $1.697 \pm 1.097$ \\

    FreeSurfer
    & --
    & $6.307 \pm 0.608$
    & --
    & $6.999 \pm 0.608$ \\

    \midrule

    SimCortex v2
    & $\underline{0.229 \pm 0.063}$
    & $\mathbf{8.81\times10^{-5} \pm 1.45\times10^{-3}}$
    & $\mathbf{0.283 \pm 0.097}$
    & $\mathbf{1.55\times10^{-4} \pm 1.10\times10^{-3}}$ \\

    \bottomrule
  \end{tabular}

  \endgroup
\end{table*}

\subsubsection{Surface-distance accuracy and self-intersection}
\label{sec:results_assd_sif}

Table~\ref{tab:dataset_method_assd_sif_means} jointly reports per-dataset ASSD and SIF. Because the FreeSurfer surfaces served as the reference for ASSD, a FreeSurfer ASSD self-comparison is not reported.

The per-dataset ASSD values reproduce those of Section~\ref{sec:results_generalization}; the self-intersection results reported alongside them show a markedly different pattern.
The FreeSurfer reference surfaces had the lowest SIF on every dataset, with dataset-level means ranging from \(0.001\%\) to \(0.008\%\). Among the learning-based methods, SimCortex v2 achieved the lowest SIF on all 14 datasets, with values ranging from \(0.022\%\) to \(0.055\%\). 
V2C-Flow, despite obtaining the lowest ASSD on most datasets, had the highest SIF among the learning-based methods on all 14 datasets, ranging from \(1.452\%\) to \(4.297\%\).

The same contrast remained after pooling by domain. 
SimCortex v2 achieved SIF values of \(0.044\%\) across the external cases and \(0.037\%\) across the held-out in-distribution cases, compared with \(2.878\%\) and \(2.392\%\), respectively, for V2C-Flow. These results indicate that low
surface-distance error does not necessarily imply low self-intersection:
SimCortex v2 maintained competitive ASSD while substantially reducing
self-intersection across both in distribution and external cohorts.

\begin{figure*}[t]
  \centering
  
  \begin{subfigure}{0.5\textwidth}
    \centering
    \includegraphics[width=\linewidth]{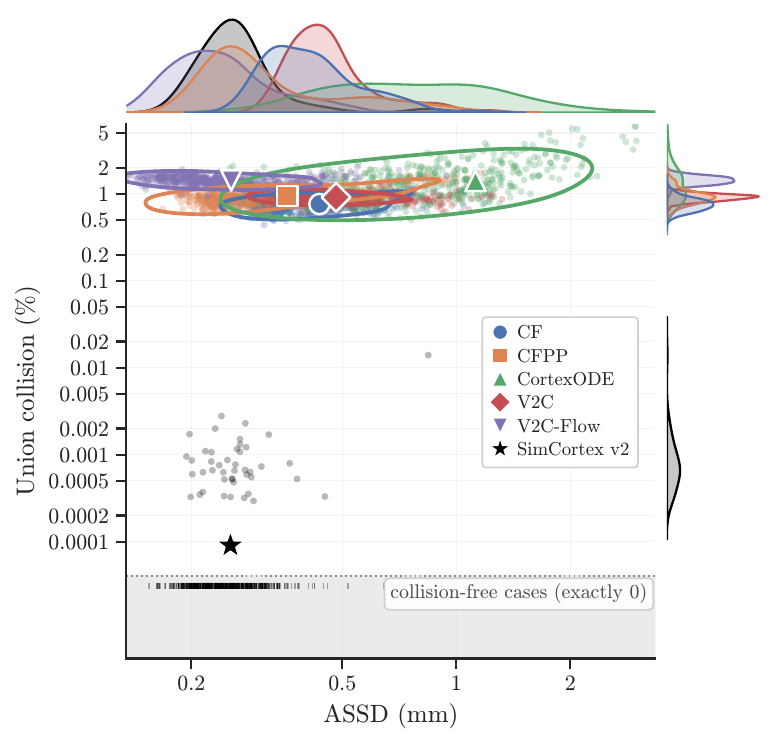}
    \caption{ASSD and Collision}
    \label{fig:assd_collision}
  \end{subfigure}\hfill
  \begin{subfigure}{0.5\textwidth}
    \centering
    \includegraphics[width=\linewidth]{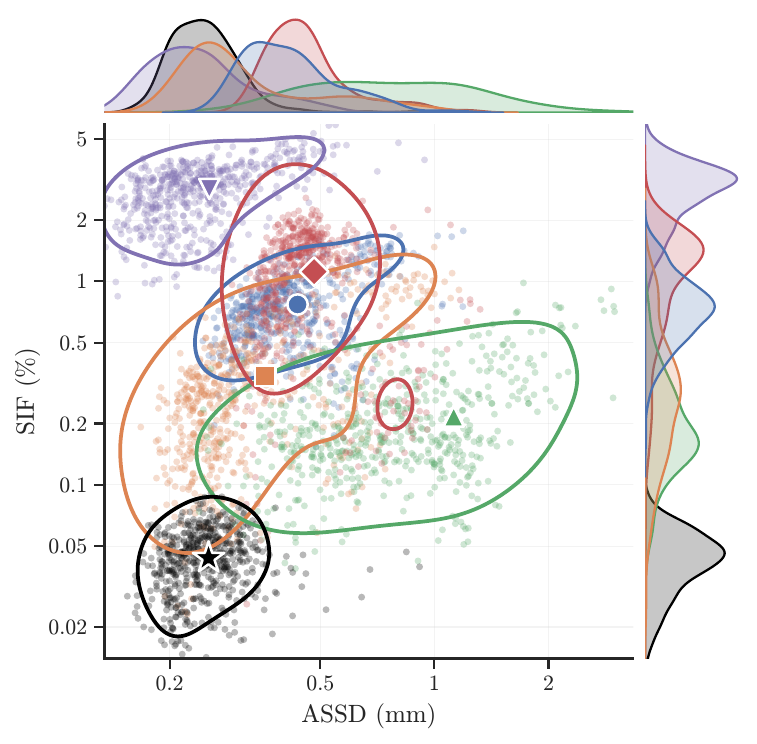}
    \caption{ASSD and SIF}
    \label{fig:assd_sif}
  \end{subfigure}
  
  \caption{Joint distributions of reconstruction accuracy and geometric
    validity across the six learning-based methods, with 560 cases per method.
    (a) ASSD versus union collision percentage. (b) ASSD versus SIF. Small markers represent individual
    cases, large markers indicate method means, and contours enclose 80\% of
    each method's cases; marginal densities are shown along the corresponding
    axes. The ASSD axis and the positive-valued collision and SIF axes are shown
    on logarithmic scales, with densities and contours fitted in log space.
    In panel~(a), union collision is exactly zero for 92.14\% of SimCortex v2
    cases and for no case of any competing method. Because zero values cannot
    be represented on a logarithmic axis, collision-free cases are shown
    separately in the shaded band at the bottom of the panel. No density contour
    is drawn for SimCortex v2 in panel~(a) because its collision distribution is
    dominated by the mass at exactly zero; for the same reason, its mean
    collision percentage lies below its smallest non-zero case. Together, the
    two panels show that similar surface-distance accuracy can be associated
    with markedly different geometric validity. Lower values are better for all
    metrics.}
  \label{fig:methods_comparison_assd_collision_sif}
\end{figure*}

Figure~\ref{fig:methods_comparison_assd_collision_sif} provides a
complementary case-level view of the relationship between reconstruction
accuracy and geometric validity. SimCortex v2 combines low ASSD with
near-zero inter-surface collision in panel~(a), including a large mass of
cases with exactly zero collision, while panel~(b) shows that it maintains
low ASSD together with substantially lower SIF than the competing
learning-based methods. These distributions illustrate that similar
surface-distance accuracy can be associated with markedly different
geometric validity.

Figures~\ref{fig:qualitative_accuracy} and~\ref{fig:qualitative_collisions} provide complementary qualitative examples for a randomly selected evaluation subject. 
Figure~\ref{fig:qualitative_accuracy} compares the reconstructed surfaces from each method with SimCortex v2 and the FreeSurfer reference on the same anatomical section, while Fig.~\ref{fig:qualitative_collisions} shows selected collision locations for each comparison method and the corresponding SimCortex v2 reconstruction at the same anatomical location.

\begin{table*}[t]
  \centering
  \caption{ASSD and SIF for each method per dataset. 
  For each case, both metrics are averaged across the four cortical surfaces and then averaged across cases within each dataset. The out of domain \textbf{OOD overall} row
  pools the 520 cases from the 13 external datasets, whereas the in distribution \textbf{ID overall} row contains the 40 held-out HCP-YA/OASIS-1 cases in
  \texttt{hcp\_oasis}. Because the FreeSurfer surfaces serve as the
  reference for ASSD, a FreeSurfer ASSD value is not reported. Lower values
  are better for both metrics. Best and second-best values are shown in bold
  and underlined, respectively.}
  \label{tab:dataset_method_assd_sif_means}

  \renewcommand{\arraystretch}{1.08}
  \setlength{\tabcolsep}{4pt}

  \begingroup
  \resulttablefont

  \begin{tabular*}{\textwidth}{
    @{\extracolsep{\fill}}
    l
    cc cc cc cc cc cc cc
    @{}
  }

    \toprule
    & \multicolumn{2}{c}{V2C\text{-}Flow}
    & \multicolumn{2}{c}{CFPP}
    & \multicolumn{2}{c}{CF}
    & \multicolumn{2}{c}{V2C}
    & \multicolumn{2}{c}{CortexODE}
    & \multicolumn{2}{c}{FreeSurfer}
    & \multicolumn{2}{c}{SimCortex v2} \\

    \cmidrule(lr){2-3}
    \cmidrule(lr){4-5}
    \cmidrule(lr){6-7}
    \cmidrule(lr){8-9}
    \cmidrule(lr){10-11}
    \cmidrule(lr){12-13}
    \cmidrule(lr){14-15}

    Dataset
    & ASSD & SIF
    & ASSD & SIF
    & ASSD & SIF
    & ASSD & SIF
    & ASSD & SIF
    & ASSD & SIF
    & ASSD & SIF \\

    \midrule

    cnp
    & $\mathbf{0.188}$ & 2.111
    & 0.244 & 0.253
    & 0.324 & 0.629
    & 0.407 & 0.662
    & 0.504 & 0.290
    & -- & $\mathbf{0.002}$
    & $\underline{0.218}$ & $\underline{0.041}$ \\

    ds000115
    & $\mathbf{0.185}$ & 2.396
    & 0.225 & 0.221
    & 0.311 & 0.651
    & 0.357 & 0.861
    & 0.490 & 0.163
    & -- & $\mathbf{0.001}$
    & $\underline{0.206}$ & $\underline{0.039}$ \\

    ds000144
    & $\mathbf{0.274}$ & 2.923
    & 0.295 & 0.384
    & 0.451 & 0.977
    & 0.513 & 1.546
    & 0.982 & 0.166
    & -- & $\mathbf{0.008}$
    & $\underline{0.282}$ & $\underline{0.054}$ \\

    ds001486
    & $\underline{0.294}$ & 3.488
    & 0.325 & 0.442
    & 0.422 & 0.824
    & 0.469 & 1.454
    & 0.929 & 0.250
    & -- & $\mathbf{0.004}$
    & $\mathbf{0.287}$ & $\underline{0.052}$ \\

    ds001748
    & $\mathbf{0.233}$ & 2.177
    & 0.253 & 0.122
    & 0.387 & 0.578
    & 0.425 & 0.890
    & 0.653 & 0.140
    & -- & $\mathbf{0.001}$
    & $\underline{0.244}$ & $\underline{0.035}$ \\

    ds002424
    & $\underline{0.313}$ & 3.492
    & 0.424 & 0.640
    & 0.451 & 0.850
    & 0.470 & 1.435
    & 1.071 & 0.288
    & -- & $\mathbf{0.004}$
    & $\mathbf{0.301}$ & $\underline{0.054}$ \\

    ds002862
    & $\underline{0.421}$ & 4.297
    & 0.826 & 0.956
    & 0.715 & 1.346
    & 0.480 & 1.510
    & 1.496 & 0.532
    & -- & $\mathbf{0.003}$
    & $\mathbf{0.284}$ & $\underline{0.035}$ \\

    ds002886
    & $\underline{0.267}$ & 3.598
    & 0.285 & 0.418
    & 0.395 & 0.844
    & 0.448 & 1.468
    & 0.958 & 0.231
    & -- & $\mathbf{0.003}$
    & $\mathbf{0.261}$ & $\underline{0.055}$ \\

    ds003499
    & $\mathbf{0.201}$ & 3.051
    & 0.263 & 0.145
    & 0.422 & 0.660
    & 0.406 & 1.323
    & 0.838 & 0.153
    & -- & $\mathbf{0.002}$
    & $\underline{0.221}$ & $\underline{0.047}$ \\

    ds003568
    & $\mathbf{0.220}$ & 2.352
    & 0.320 & 0.268
    & 0.406 & 0.529
    & 0.729 & 0.621
    & 0.950 & 0.196
    & -- & $\mathbf{0.002}$
    & $\underline{0.252}$ & $\underline{0.053}$ \\

    ds003763
    & $\mathbf{0.199}$ & 2.768
    & 0.295 & 0.167
    & 0.357 & 0.629
    & 0.388 & 0.963
    & 0.883 & 0.190
    & -- & $\mathbf{0.003}$
    & $\underline{0.225}$ & $\underline{0.022}$ \\

    ds005234
    & $\mathbf{0.263}$ & 3.306
    & 0.328 & 0.371
    & 0.417 & 0.949
    & 0.478 & 1.844
    & 0.735 & 0.199
    & -- & $\mathbf{0.008}$
    & $\underline{0.264}$ & $\underline{0.048}$ \\

    ds006067
    & $\mathbf{0.239}$ & 1.452
    & 0.574 & 0.176
    & 0.568 & 0.459
    & 0.763 & 0.218
    & 1.208 & 0.101
    & -- & $\mathbf{0.002}$
    & $\underline{0.295}$ & $\underline{0.039}$ \\

    \midrule

    \textbf{OOD} ($n=520$)
    & $\mathbf{0.254}$ & 2.878
    & 0.358 & 0.351
    & 0.433 & 0.764
    & 0.487 & 1.138
    & 0.900 & 0.223
    & -- & $\mathbf{0.003}$
    & $\underline{0.257}$ & $\underline{0.044}$ \\

    \textbf{ID} ($n=40$)
    & $\underline{0.268}$ & 2.392
    & 0.353 & 0.241
    & 0.469 & 0.849
    & 0.410 & 0.824
    & 4.041 & 0.137
    & -- & $\mathbf{0.003}$
    & $\mathbf{0.207}$ & $\underline{0.037}$ \\

    \bottomrule

  \end{tabular*}

  \endgroup
\end{table*}

\begin{figure*}[t]
    \centering
    \setlength{\tabcolsep}{2pt}

    \begin{tabular}{
        @{}
        *{5}{>{\centering\arraybackslash}m{0.19\textwidth}}
        @{}
    }

    \figlabel{V2C-Flow} &
    \figlabel{CFPP} &
    \figlabel{CF} &
    \figlabel{V2C} &
    \figlabel{CortexODE}
    \\[1mm]

    \includegraphics[width=\linewidth]{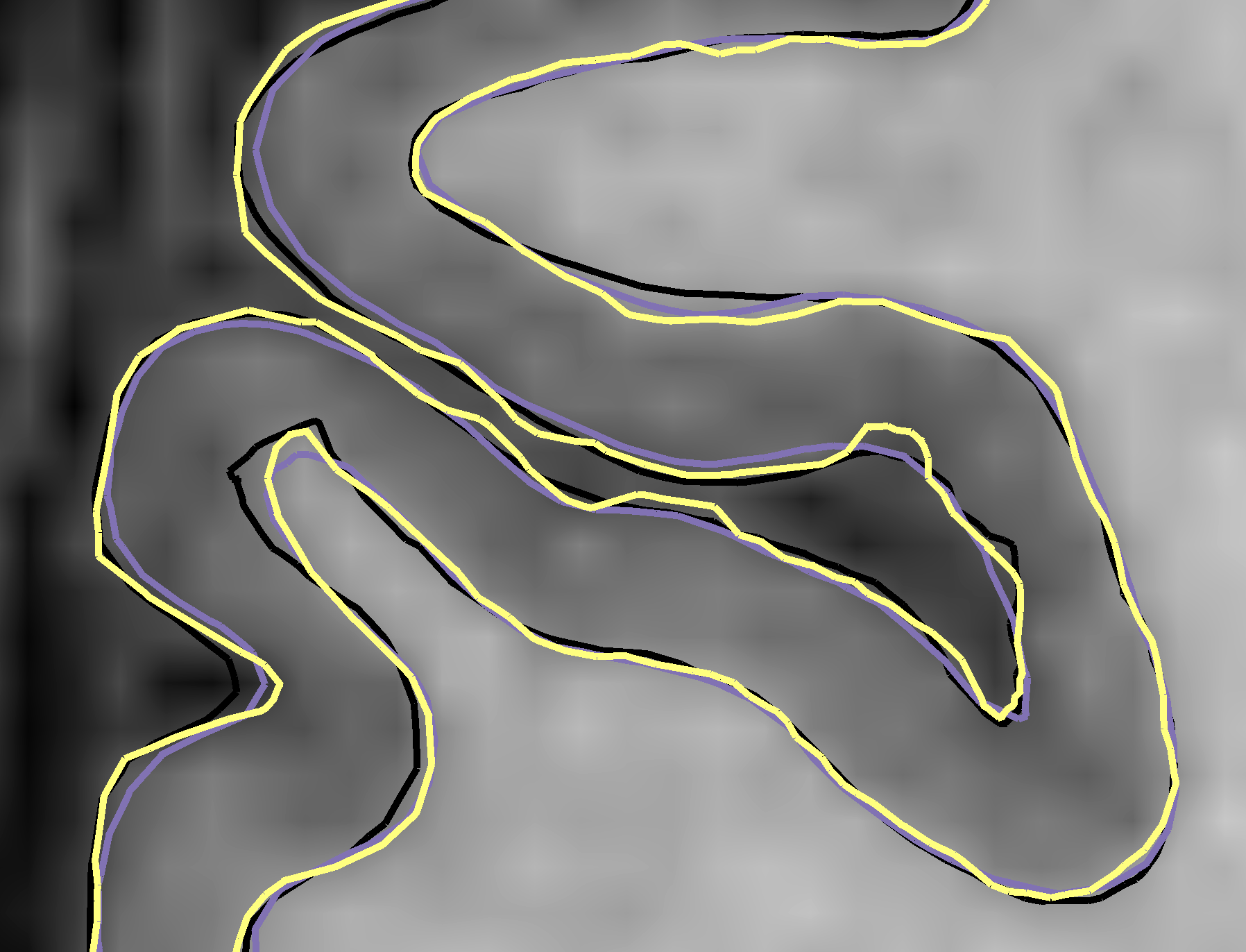} &
    \includegraphics[width=\linewidth]{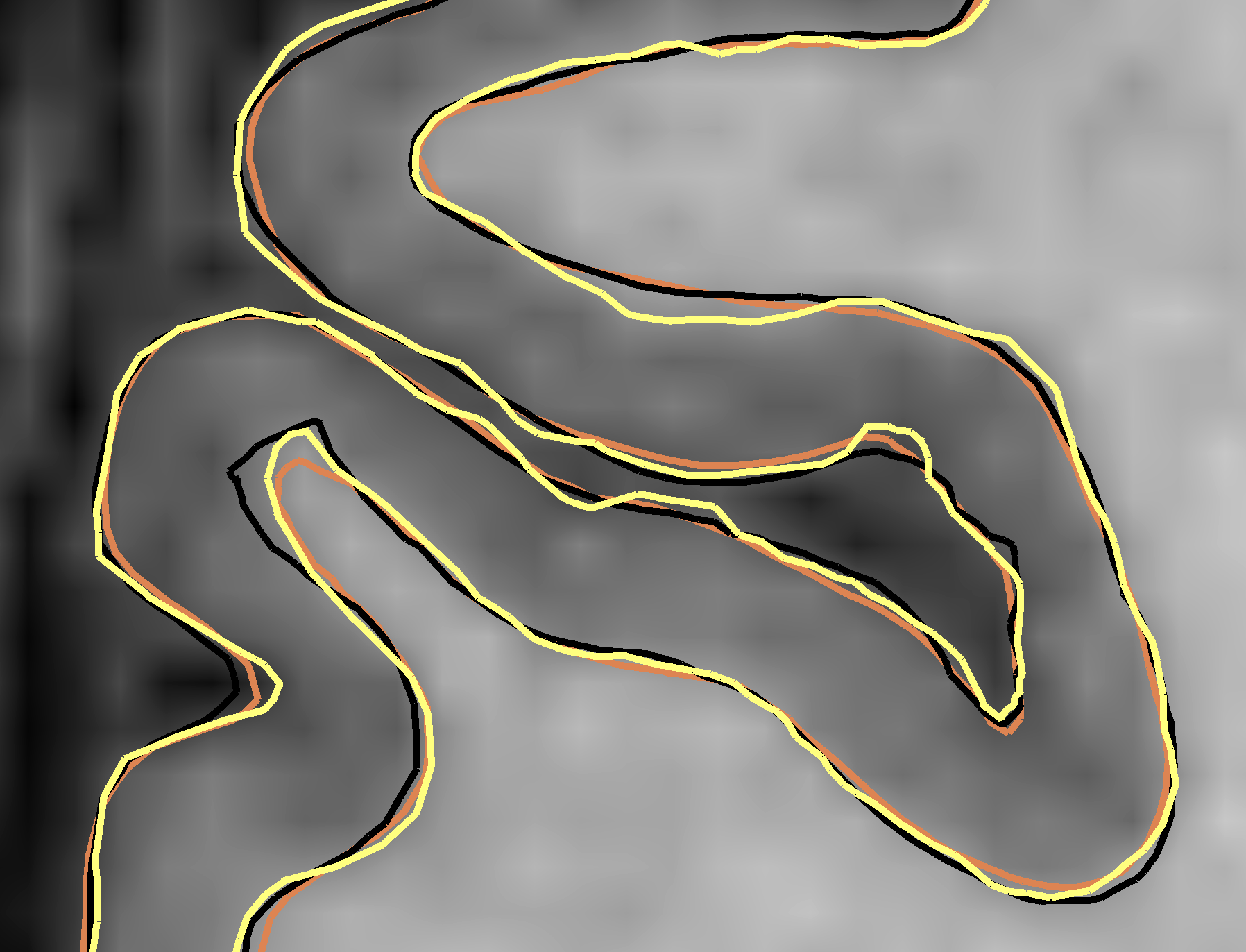} &
    \includegraphics[width=\linewidth]{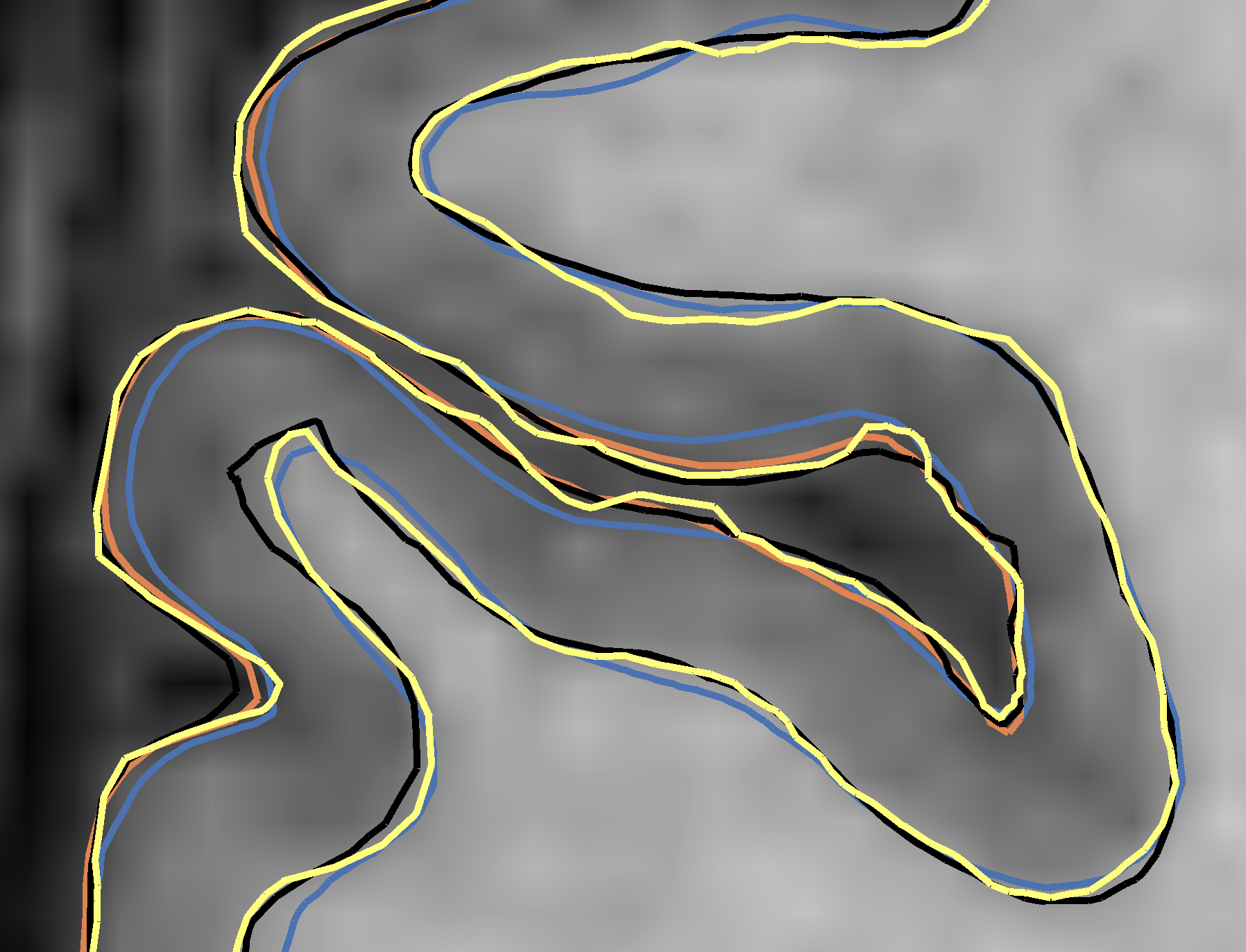} &
    \includegraphics[width=\linewidth]{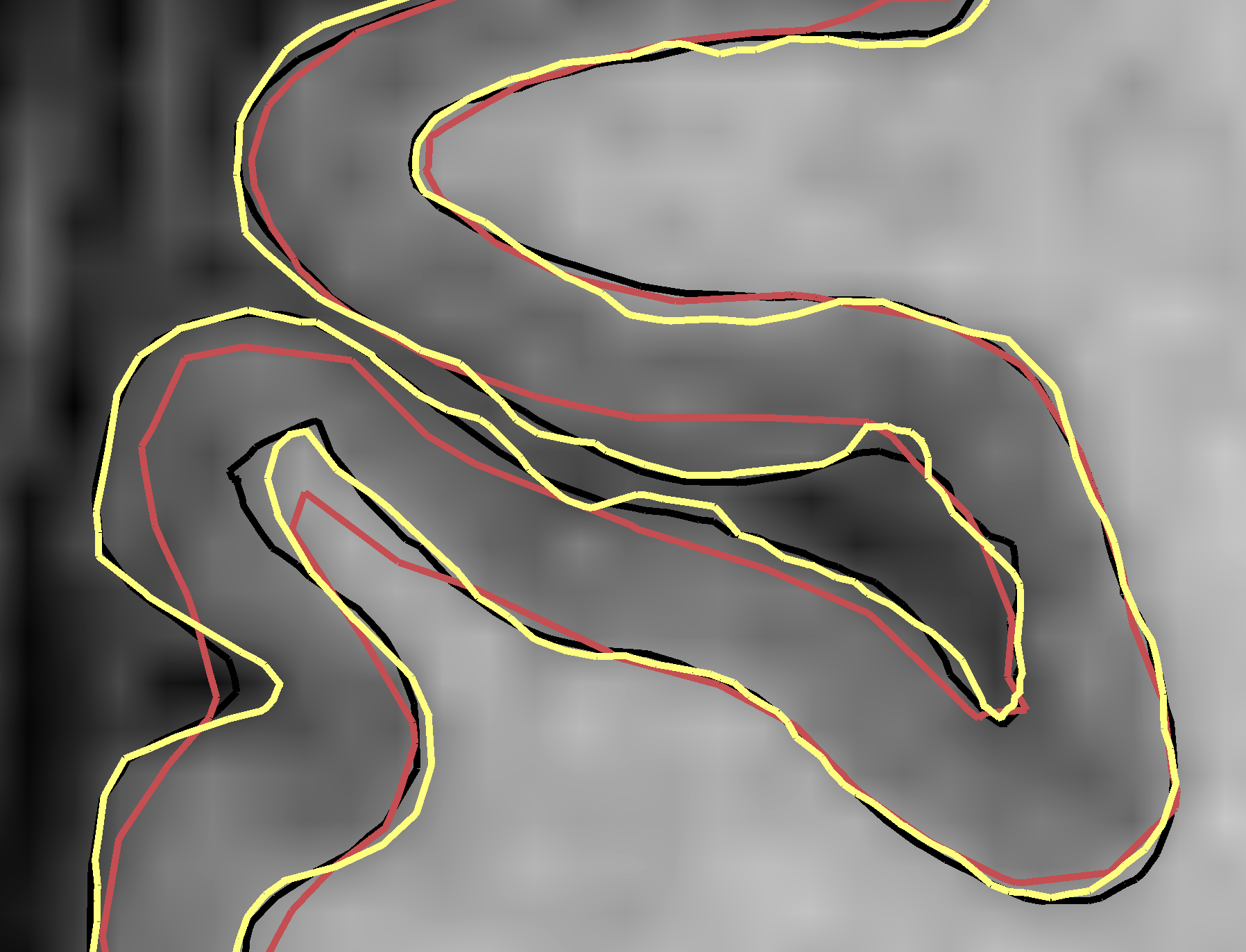} &
    \includegraphics[width=\linewidth]{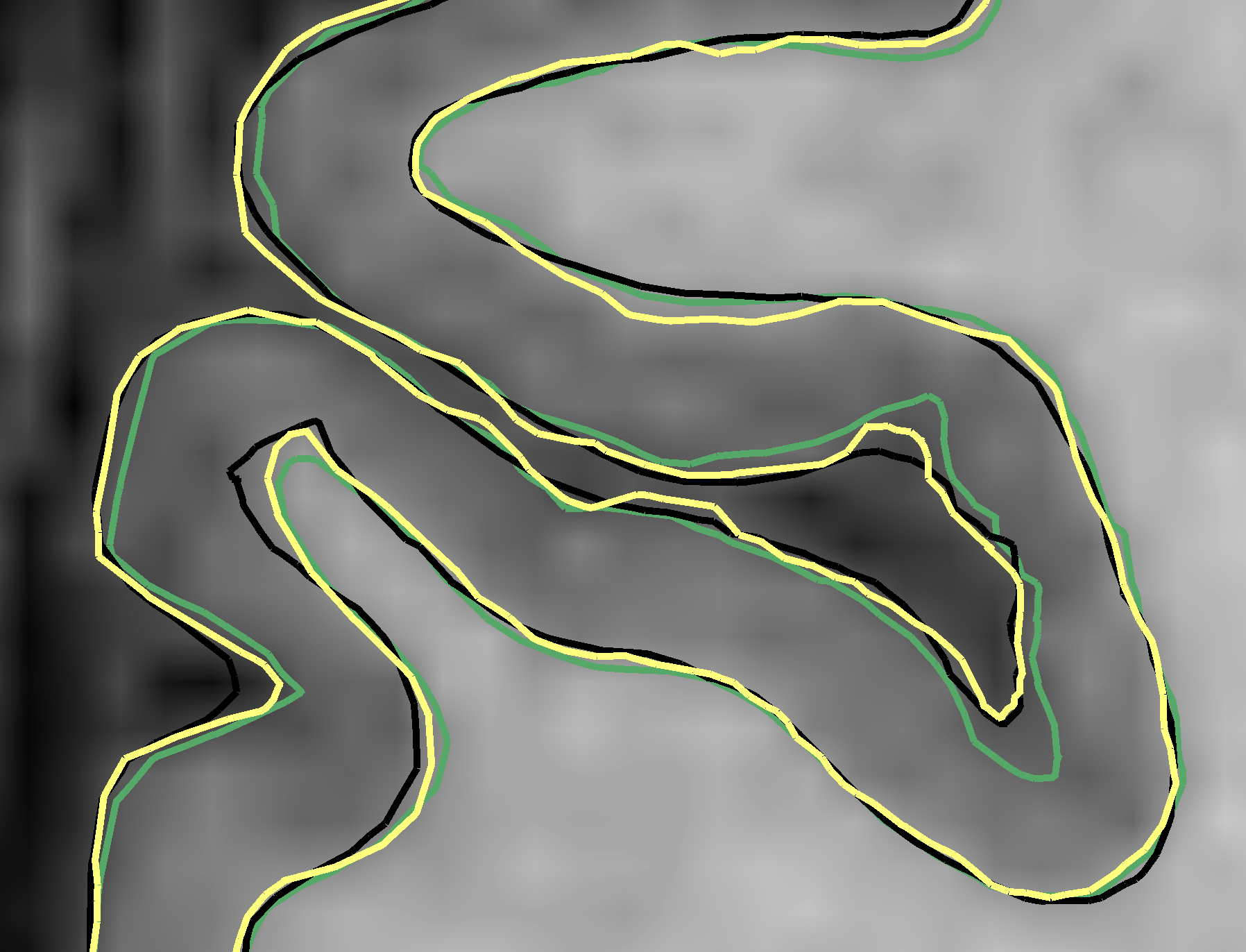}

    \end{tabular}

\caption{
Qualitative comparison of cortical surface reconstruction in a complex,
highly folded cortical region for a randomly selected evaluation subject.
The colored contours show the pial and white cortical surfaces on the displayed MRI slice.
In every panel, SimCortex v2 is shown in black and the FreeSurfer reference in
yellow; the remaining colored contour corresponds to the comparison method
named above the panel.
All panels show the same anatomical location and field of view, allowing
visual comparison of surface localization in deep sulcal regions.
}

    \label{fig:qualitative_accuracy}
\end{figure*}

\begin{figure*}[t]
    \centering
    \setlength{\tabcolsep}{0.5pt}
    \renewcommand{\arraystretch}{0.95}

    \begin{tabular}{
        @{}
        >{\centering\arraybackslash}m{0.025\textwidth}
        *{6}{>{\centering\arraybackslash}m{0.159\textwidth}}
        @{}
    }

    &
    \figlabel{V2C-Flow} &
    \figlabel{CFPP} &
    \figlabel{CF} &
    \figlabel{V2C} &
    \figlabel{CortexODE} &
    \figlabel{FreeSurfer}
    \\[1mm]

    &
    \includegraphics[width=\linewidth]{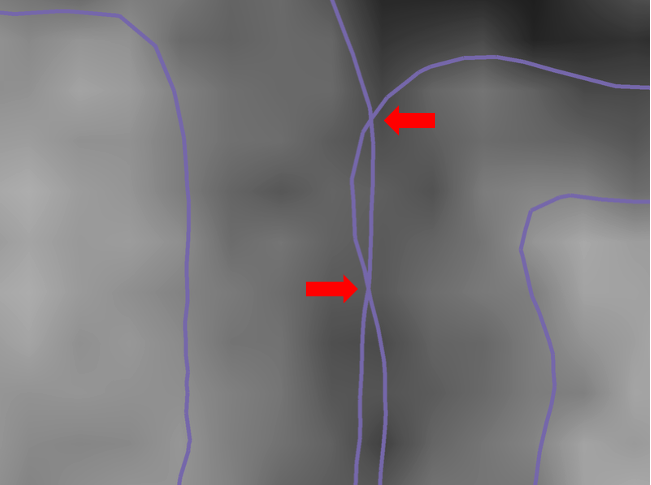} &
    \includegraphics[width=\linewidth]{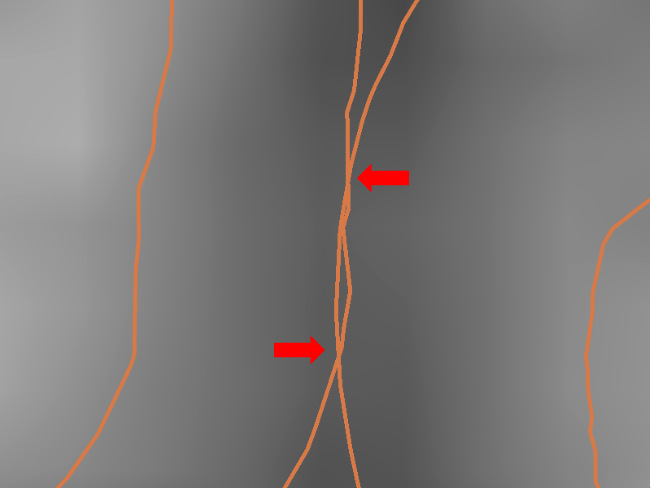} &
    \includegraphics[width=\linewidth]{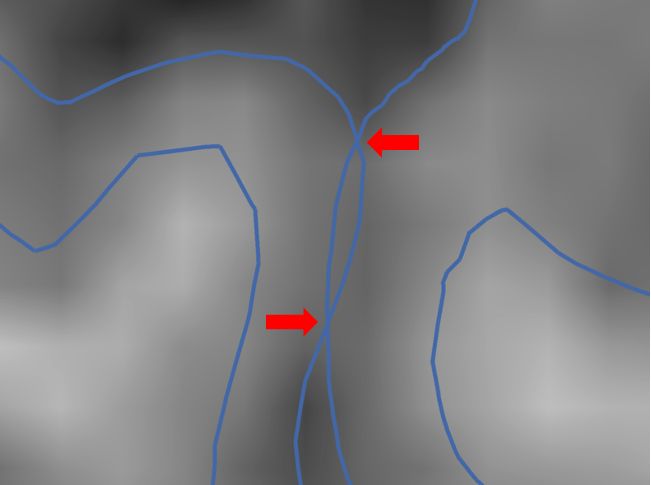} &
    \includegraphics[width=\linewidth]{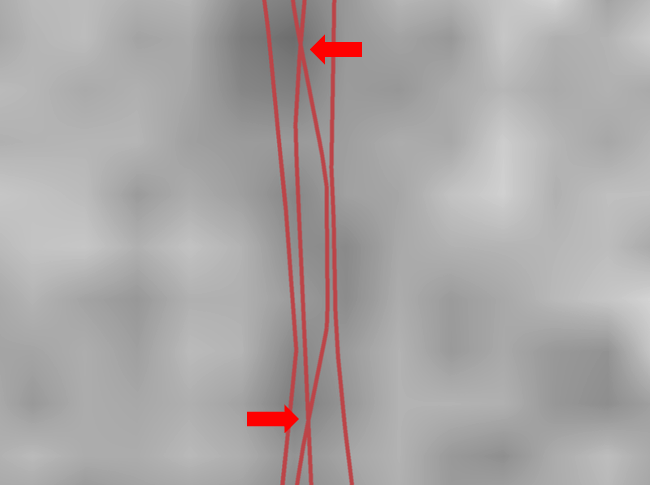} &
    \includegraphics[width=\linewidth]{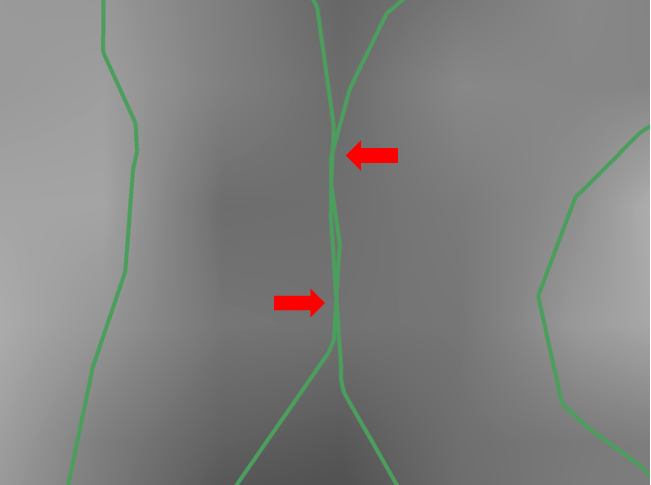} &
    \includegraphics[width=\linewidth]{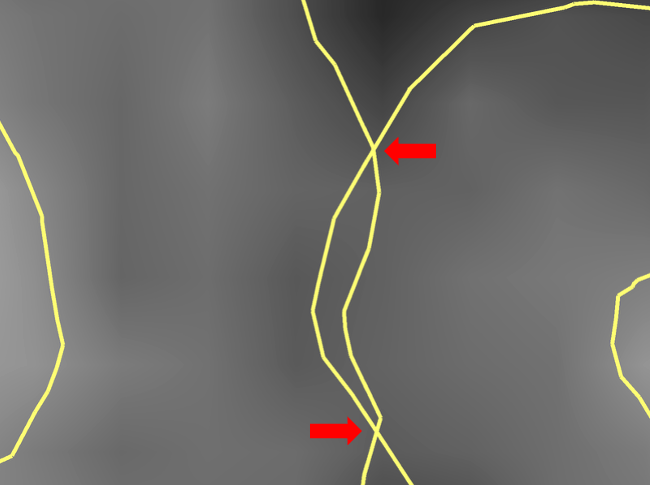}
    \\[1.5mm]

    \raisebox{6mm}{%
        \rotatebox[origin=c]{90}{\figlabel{SimCortex}}
    } &
    \includegraphics[width=\linewidth]{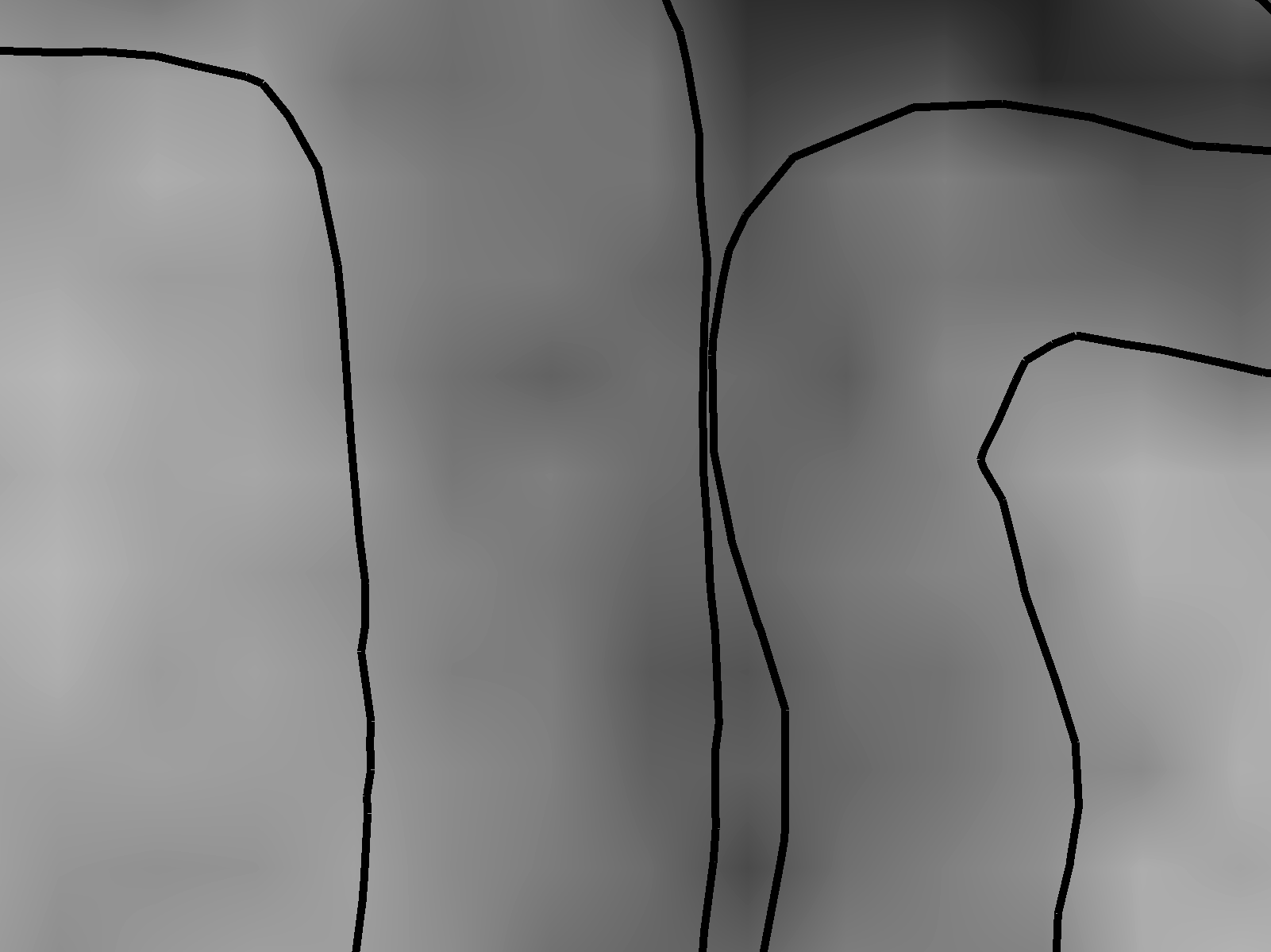} &
    \includegraphics[width=\linewidth]{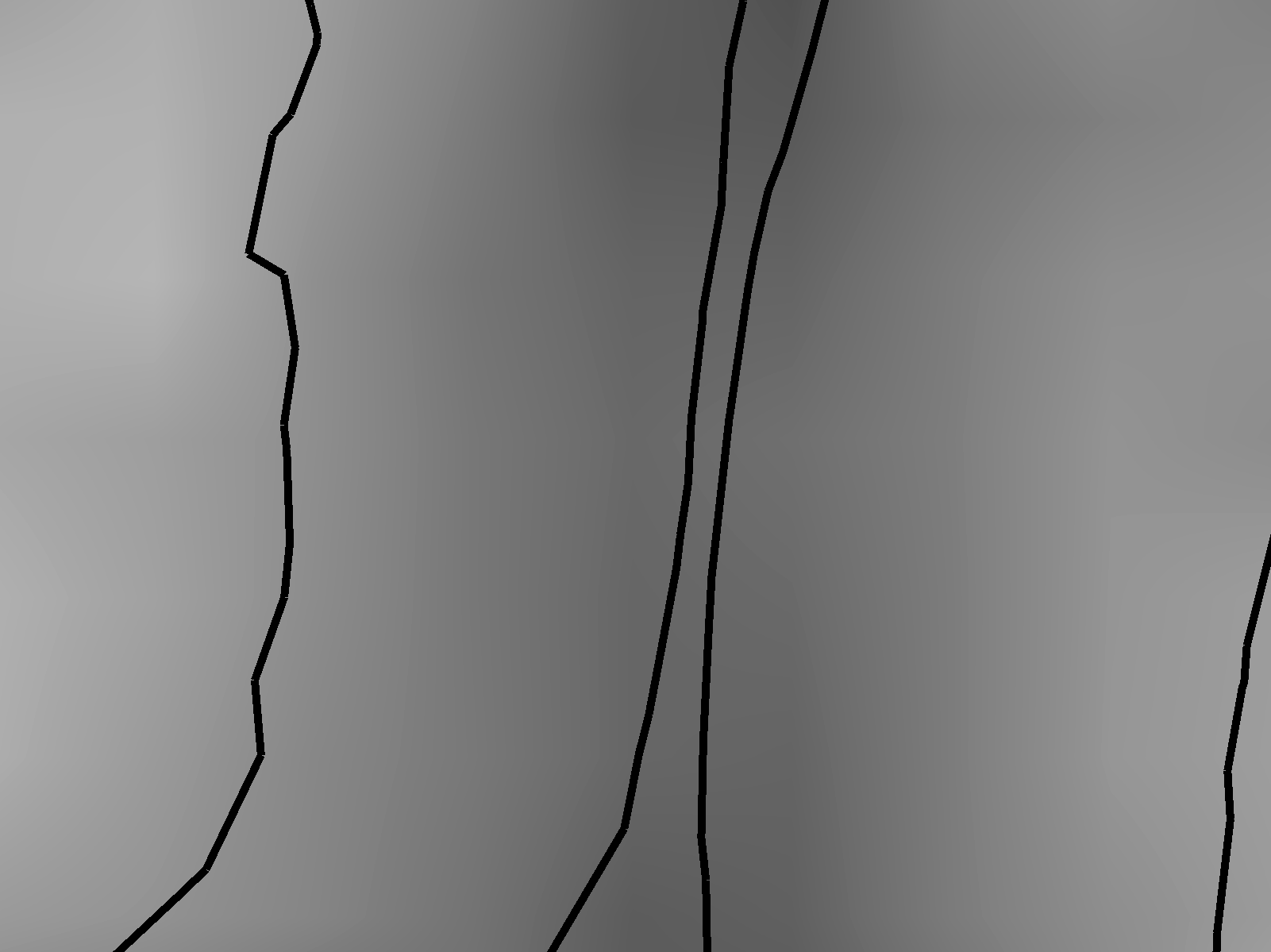} &
    \includegraphics[width=\linewidth]{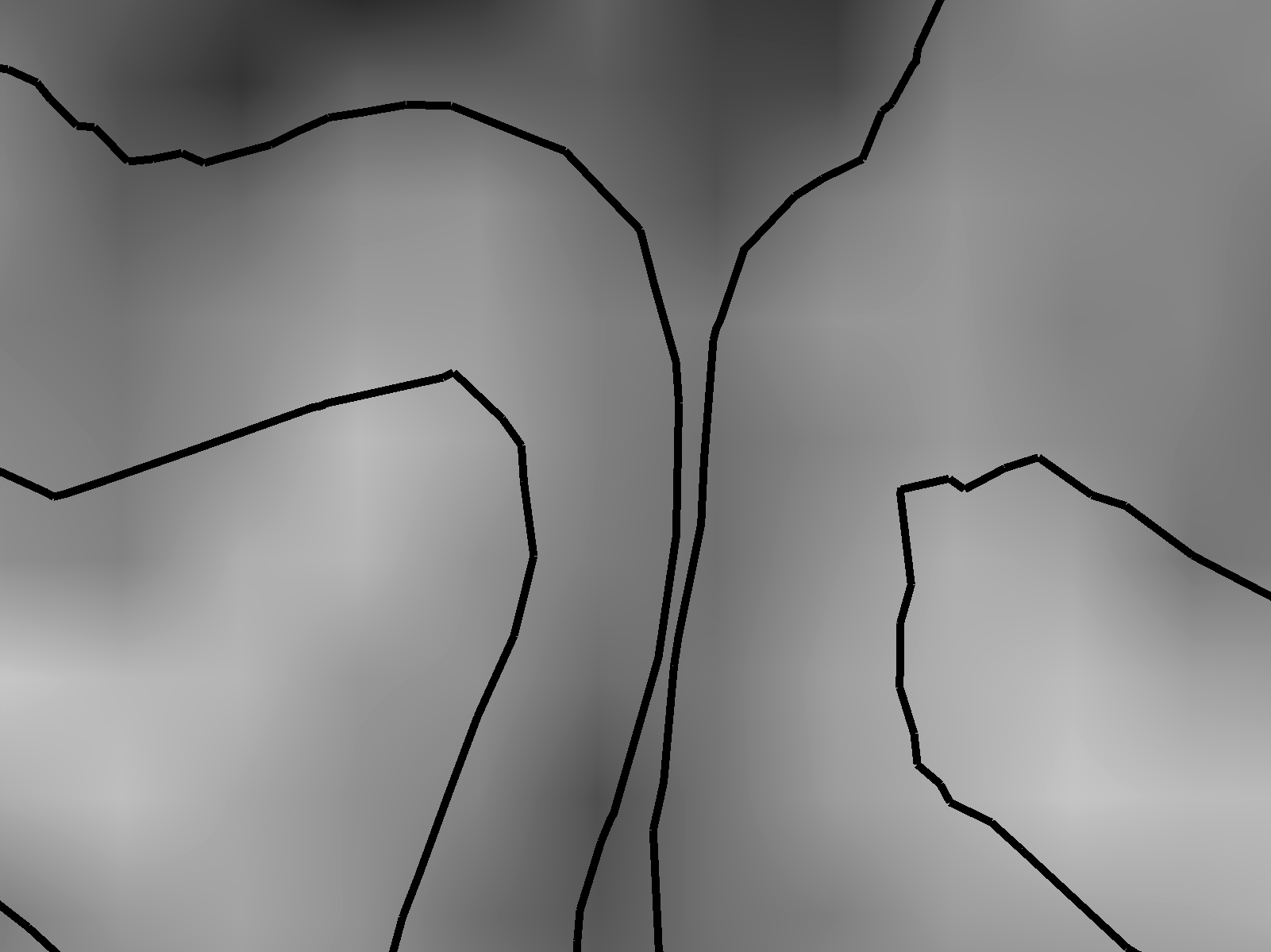} &
    \includegraphics[width=\linewidth]{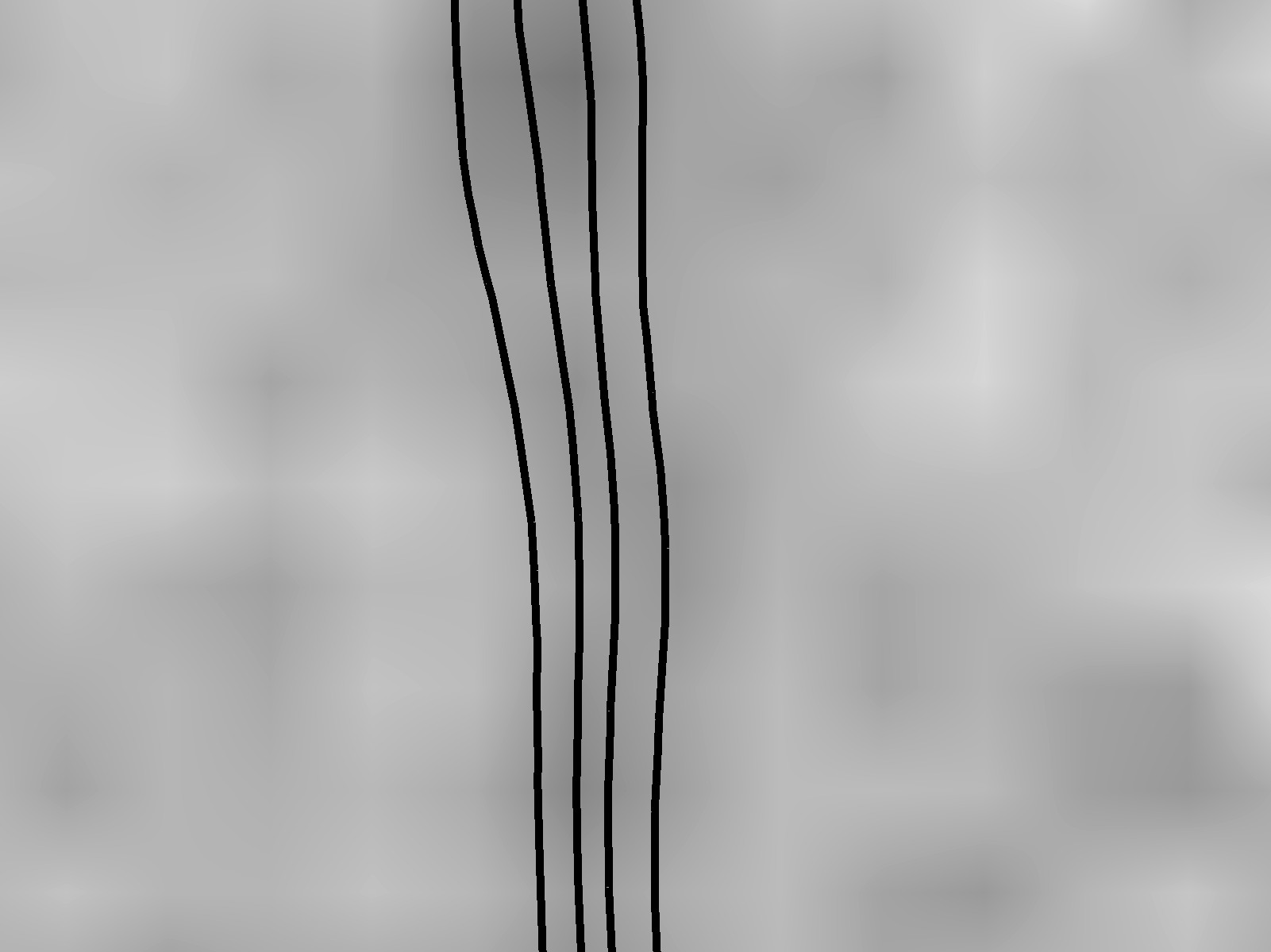} &
    \includegraphics[width=\linewidth]{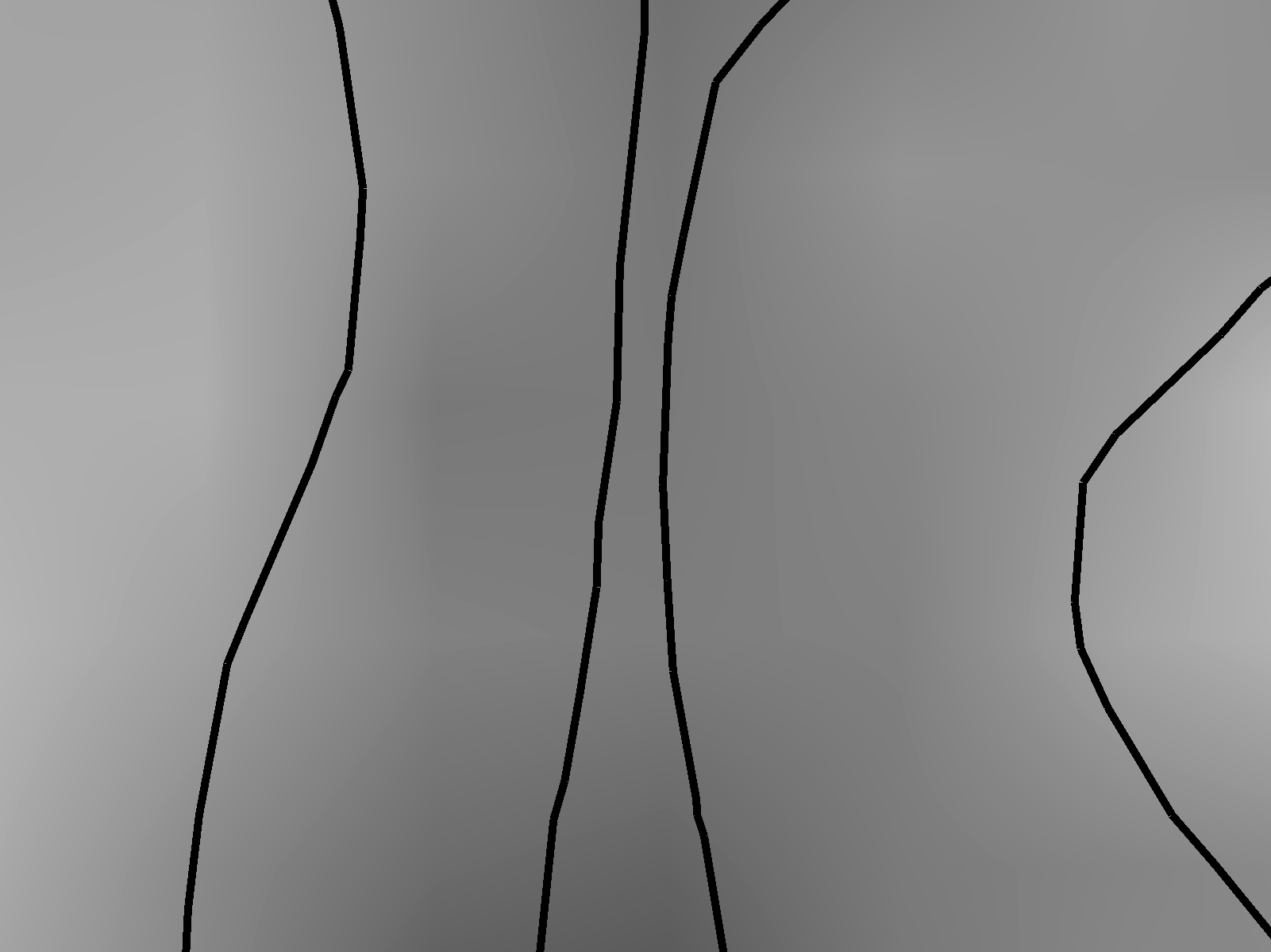} &
    \includegraphics[width=\linewidth]{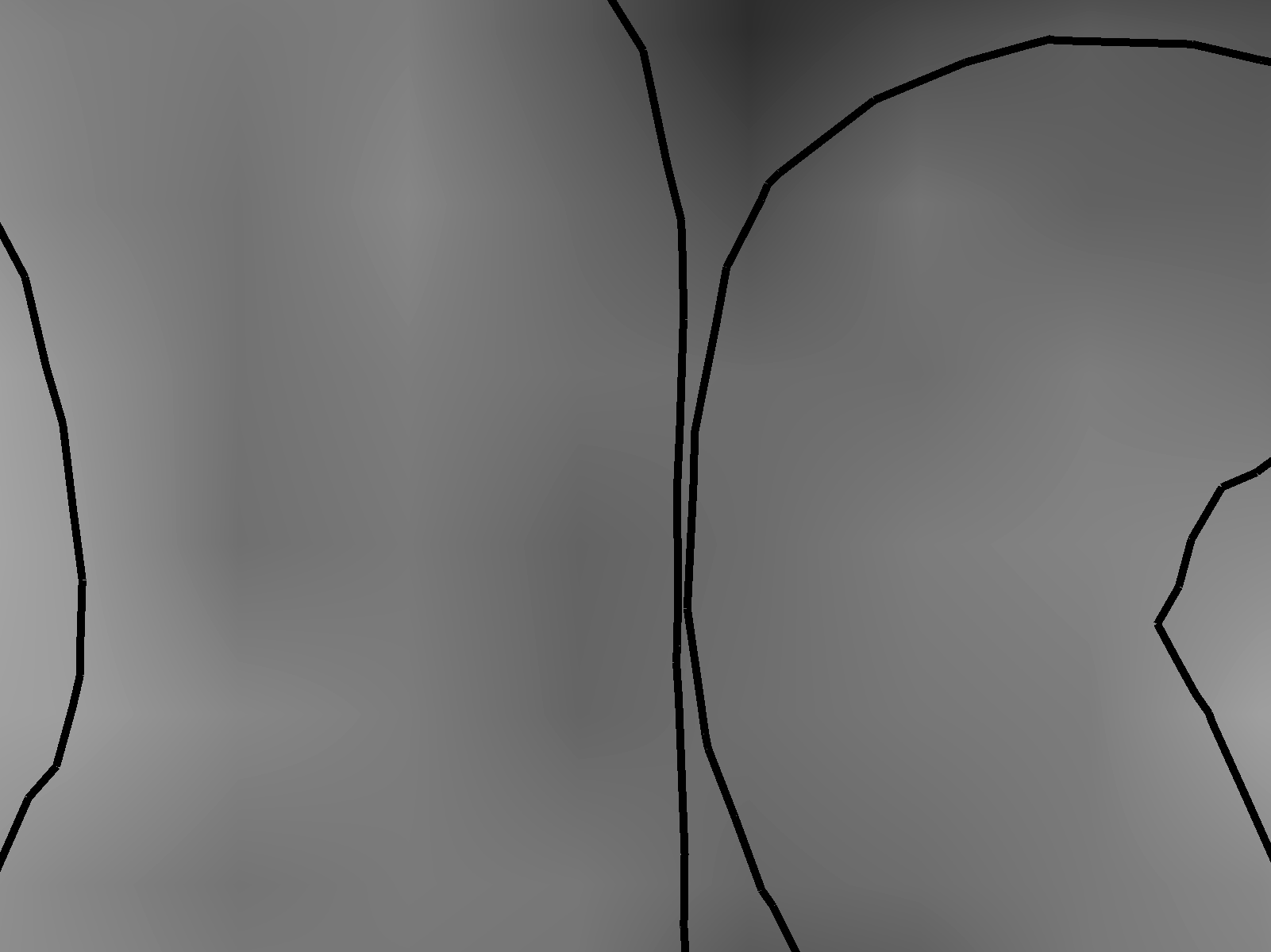}
    \\

    \end{tabular}

    \caption{
    Qualitative examples of left--right pial-surface collisions for the same
    randomly selected evaluation subject.
    Each top-row panel shows the pial and white surface contours produced by the
    indicated method, zoomed around a location where the left and right pial
    surfaces intersect; red arrows indicate the selected collision points.
    The bottom row shows the corresponding SimCortex v2 reconstruction at the exact
    same anatomical location and field of view.
    In the SimCortex v2 reconstructions, the corresponding left and right pial
    surfaces remain separated.
    }

    \label{fig:qualitative_collisions}
\end{figure*}

\subsection{Ablation study}
\label{sec:ablation}

We evaluated five controlled variants of the deformation model, each removing
one or more components from the full model, to separate the contributions of the ribbon
representation, encoder design, ribbon--MRI fusion strategy, geometry-encoder
depth, and geometric loss terms. All variants used the same training and
evaluation cohorts, initialized surfaces, optimization protocol,
checkpoint-selection criterion, and 560-case evaluation set. Removing the
dedicated ribbon encoder and gated fusion leaves the early-fusion residual
variant, in which the T1w image and ribbon map enter a single encoder by
channel concatenation; additionally removing the residual, group-normalized
backbone leaves the Vanilla U-Net. The MRI-only variant removes the ribbon
input altogether, and with it the ribbon encoder and fusion pathway. The
shallow dual-encoder variant retains both encoders and the gated fusion but
reduces the geometry-encoder depth from six to three. The loss ablation
retains the complete architecture while setting the white--pial separation,
signed nesting, and left--right pial separation weights to zero.
Table~\ref{tab:ablation_overall} reports each variant against the full model.

\begin{figure*}[t]
  \centering
  \includegraphics[width=\textwidth]{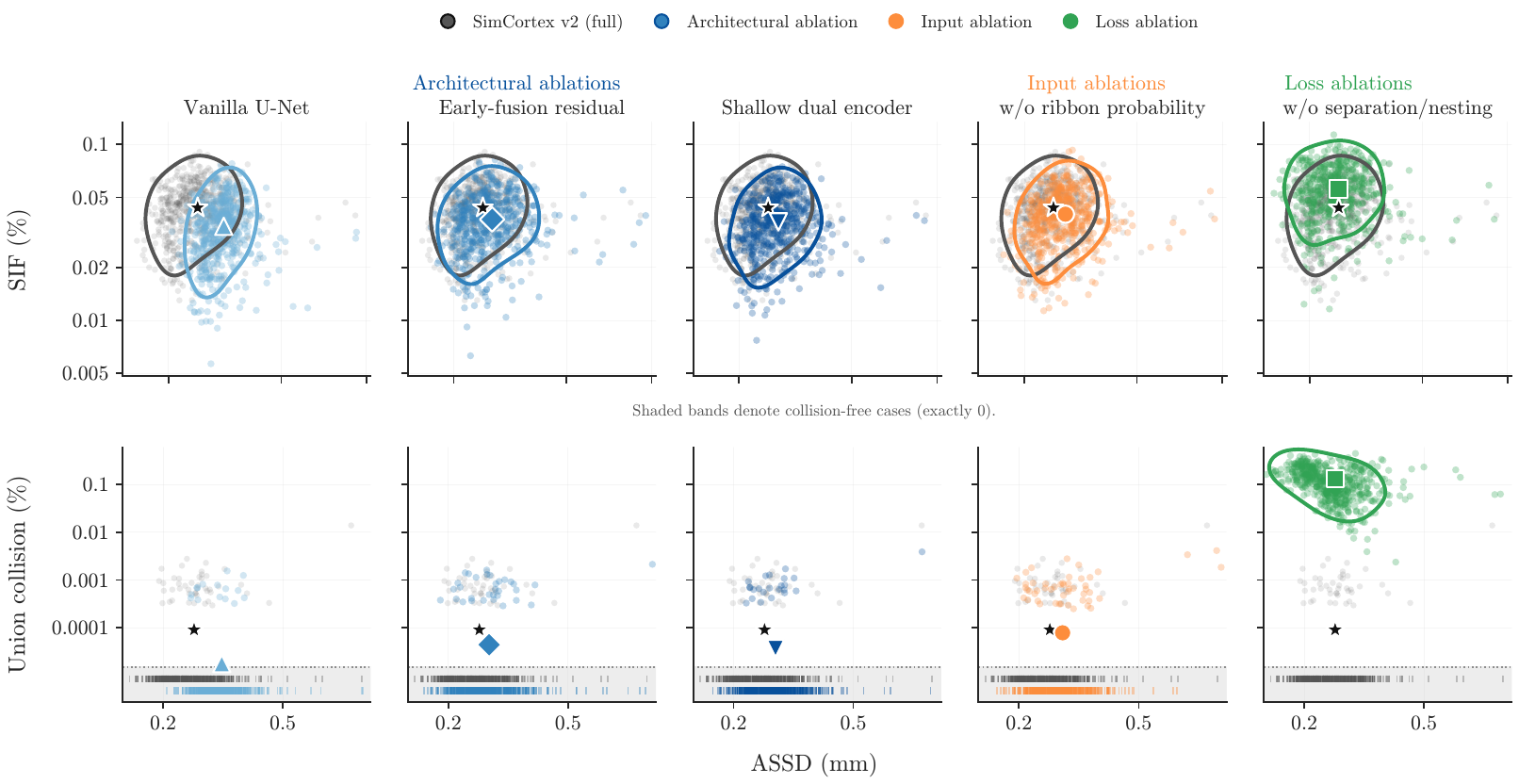}

  \caption{Case-level relationship between reconstruction accuracy and
  geometric validity for the SimCortex v2 ablation variants across the
  560 evaluation cases. Columns compare the full model with, from left to
  right, the Vanilla U-Net, early-fusion residual, shallow dual-encoder,
  MRI-only (without ribbon probability), and no-separation/nesting variants.
  The top row shows ASSD versus self-intersecting-face percentage (SIF), and
  the bottom row shows ASSD versus union inter-surface collision
  percentage. Gray points represent SimCortex v2 and colored points represent
  the corresponding ablated variant; large markers indicate arithmetic means.
  In the SIF panels, black and colored contours enclose 80\% of the cases for
  the full and ablated models, respectively. ASSD, SIF, and positive-valued
  collision measurements are displayed on logarithmic scales. Because exact
  zero collision cannot be represented on a logarithmic axis, collision-free
  cases are shown separately in the shaded bands in the bottom row. The first
  three columns group architectural ablations, whereas the final two columns
  examine removal of the ribbon input and the separation/nesting losses.
  Lower values are better for all displayed metrics.}
  \label{fig:ablation_accuracy_validity}
\end{figure*}

\paragraph{Removing the separation and nesting losses}
This removal costs essentially nothing in surface-distance accuracy and costs
the collision result entirely. ASSD is unchanged at \(0.253\,\mathrm{mm}\) and
Chamfer-\(L_1\) differs by \(0.001\,\mathrm{mm}\), while HD90 rises only from
\(0.586\) to \(0.593\,\mathrm{mm}\). Geometric validity, by contrast,
collapses: mean union collision rises from \(9.04\times10^{-5}\%\) to
\(0.132\%\), a factor of roughly \(1500\), and the proportion of completely
collision-free cases falls from \(92.14\%\) to zero. SIF also rises, from
\(0.044\%\) to \(0.056\%\). These losses therefore buy the collision-free
rate outright, at no measurable cost in mean surface localization. The
contrast is visible in the rightmost column of
Fig.~\ref{fig:ablation_accuracy_validity}, where the ablated ASSD
distribution overlaps that of the full model while the collision distribution
shifts upward by several orders of magnitude and the collision-free band
empties.

\paragraph{Removing architectural components}
Each architectural removal costs surface-distance accuracy. Removing the
dedicated ribbon encoder and gated fusion increases ASSD by
\(0.020\,\mathrm{mm}\) (\(7.9\%\)) and HD90 by \(0.053\,\mathrm{mm}\), with
the full model more accurate on every one of the 14 evaluation datasets.
Additionally removing the residual, group-normalized backbone increases ASSD
by \(0.060\,\mathrm{mm}\) (\(23.7\%\)), the largest accuracy loss of any
variant. Reducing the geometry-encoder depth from six to three, while
retaining both encoders and the gated fusion, still costs
\(0.022\,\mathrm{mm}\) ASSD (\(8.7\%\)) and \(0.062\,\mathrm{mm}\) HD90, with
the full model ahead on all 14 datasets for all three surface-distance
metrics. 

\paragraph{Removing ribbon input}
Removing the ribbon input altogether costs \(0.026\,\mathrm{mm}\) ASSD (\(10.3\%\)), with the ribbon-conditioned model ahead on 13 of the 14 datasets. 
The benefit of ribbon conditioning is therefore not simply that of an additional input channel: a sufficiently deep, dedicated geometry pathway with multi-scale fusion is needed to exploit it.

\paragraph{An accuracy--validity tradeoff among architectures}
The architectural removals move accuracy and geometric validity in opposite
directions. In general, variants that lose accuracy also collide less: the Vanilla
U-Net has the worst ASSD of any variant and the lowest collision rate
(\(2.68\%\) of cases, against \(7.86\%\) for the full model), while the
shallow dual-encoder and early-fusion variants fall between the two on both
measures. This ordering is expected. Initialization is collision-free in
every case (Section~\ref{sec:surface_initialization}), so collisions can only
arise during deformation, and a variant that fits the target surfaces less
closely deforms less and departs less from that collision-free starting
configuration. Considered in isolation, the collision rate therefore rewards
underfitting, which is why the full model does not attain the lowest value
among the architectural variants.

The separation and nesting losses are what break this tradeoff. They allow
the full model to sit at the accurate end of the range while retaining a
collision rate of \(7.86\%\), against \(100\%\) for the same architecture
trained without them. The full model therefore combines the best
surface-distance accuracy of any variant with a collision rate comparable to
that of variants far less accurate, and with near-zero mean union collision.

\begin{table*}[t]
  \centering
  \caption{Ablation results across the 560 evaluation cases. The full model is
  the reference; each block below removes one or more component from it and reports the
  resulting change ($\Delta$, ablated minus full). Values are mean $\pm$ SD
  across cases, except the collision rate. Lower values are better for all
  metrics, so a positive $\Delta$ indicates a loss. Deltas are computed from
  unrounded values. Because the variants are diagnostic probes rather than
  competing methods, no best-in-column marking is used; the accuracy--validity
  tradeoff visible among the architectural variants is discussed in
  Section~\ref{sec:ablation}.}
  \label{tab:ablation_overall}

  \renewcommand{\arraystretch}{1.08}
  \setlength{\tabcolsep}{3pt}

  \begingroup
  \resulttablefont

  \begin{tabular*}{\textwidth}{
    @{\extracolsep{\fill}}
    l
    c c c c c c
    @{}
  }
    \toprule

    Variant
    & ASSD (mm)
    & HD90 (mm)
    & Chamfer (mm)
    & SIF (\%)
    & Union collision (\%)
    & Coll.\ (\%) \\

    \midrule

    \textbf{SimCortex v2 (full)}
    & $0.253 \pm 0.067$
    & $0.586 \pm 0.228$
    & $0.529 \pm 0.060$
    & $0.044 \pm 0.014$
    & $9.04\times10^{-5} \pm 6.47\times10^{-4}$
    & $7.86$ \\

    \midrule
    \multicolumn{7}{@{}l}{\emph{Remove architectural components}}\\

    \quad Dedicated ribbon encoder and gated fusion
    & $0.273 \pm 0.079$
    & $0.639 \pm 0.253$
    & $0.541 \pm 0.069$
    & $0.038 \pm 0.013$
    & $4.41\times10^{-5} \pm 2.08\times10^{-4}$
    & $5.71$ \\

    \quad $\Delta$
    & $+0.020$
    & $+0.053$
    & $+0.012$
    & $-0.006$
    & $-4.63\times10^{-5}$
    & $-2.15$ \\[2pt]

    \quad Residual, normalized backbone
    & $0.313 \pm 0.062$
    & $0.715 \pm 0.211$
    & $0.568 \pm 0.058$
    & $0.034 \pm 0.013$
    & $1.77\times10^{-5} \pm 1.20\times10^{-4}$
    & $2.68$ \\

    \quad $\Delta$
    & $+0.060$
    & $+0.129$
    & $+0.039$
    & $-0.010$
    & $-7.27\times10^{-5}$
    & $-5.18$ \\[2pt]

    \quad Geometry-encoder depth, six to three
    & $0.275 \pm 0.066$
    & $0.648 \pm 0.217$
    & $0.542 \pm 0.059$
    & $0.036 \pm 0.012$
    & $3.67\times10^{-5} \pm 2.30\times10^{-4}$
    & $4.11$ \\

    \quad $\Delta$
    & $+0.022$
    & $+0.062$
    & $+0.013$
    & $-0.008$
    & $-5.37\times10^{-5}$
    & $-3.75$ \\
    
    \midrule
    \multicolumn{7}{@{}l}{\emph{Remove ribbon input entirely}}\\
    \quad value
    & $0.279 \pm 0.071$
    & $0.664 \pm 0.231$
    & $0.547 \pm 0.064$
    & $0.040 \pm 0.014$
    & $7.82\times10^{-5} \pm 3.43\times10^{-4}$
    & $8.75$ \\

    \quad $\Delta$
    & $+0.026$
    & $+0.078$
    & $+0.018$
    & $-0.004$
    & $-1.22\times10^{-5}$
    & $+0.89$ \\

  \midrule
    \multicolumn{7}{@{}l}{\emph{Remove the separation and nesting losses}}\\

    \quad value
    & $0.253 \pm 0.067$
    & $0.593 \pm 0.227$
    & $0.528 \pm 0.060$
    & $0.056 \pm 0.017$
    & $0.132 \pm 0.086$
    & $100.00$ \\

    \quad $\Delta$
    & $-0.000$
    & $+0.007$
    & $-0.001$
    & $+0.012$
    & $+0.132$
    & $+92.14$ \\

    \bottomrule
  \end{tabular*}

  \endgroup
\end{table*}

\section{Discussion}
\label{sec:discussion}

SimCortex v2 was designed to reconstruct all four cortical surfaces in a single whole-brain formulation while keeping inter-surface collisions and mesh self-intersections low. 
Across 560 cases from 14 cohorts, its surface-distance accuracy was on par with the strongest baseline, V2C-Flow. 
The clearer separation appeared in geometric validity. 
SimCortex v2 was free of inter-surface collisions in $92.14\%$ of cases and had a near-zero mean union collision several orders of magnitude below every other method, whereas all baselines produced at least one collision in every case. 
It also had the lowest self-intersection fraction among the learning-based methods. The main advantage of the method is therefore not a large reduction in surface-distance error, but the ability to reach comparable accuracy while almost entirely avoiding the anatomically implausible surface interactions that the other methods exhibit.

The comparison also shows that surface-distance accuracy, self-intersection, and inter-surface collision measure different properties and do not improve together.
V2C-Flow reached the lowest cortical thickness error but had the highest SIF and union collision among the learning-based methods, indicating accurate boundary placement accompanied by frequent geometric defects.
FreeSurfer showed the opposite pattern, with the lowest SIF but the highest collision of any method. 
SIF and inter-surface collision capture, respectively, self-intersection within a single surface and overlap between
adjacent surfaces, and a method can perform well on one while performing poorly on the other. 
Reporting surface distance, SIF, and collision separately, rather than a single accuracy score, is thus necessary to characterize reconstruction quality, and SimCortex v2 is the only evaluated method designed to be simultaneously competitive on all three.

These findings were consistent across cohorts. 
SimCortex v2 ranked first or second in ASSD on every one of the 14 datasets and had the lowest union collision on every dataset. 
The near-zero overall collision was reflected at the level of individual datasets rather than produced by a subset of cases, which suggests that the collision behavior follows from the simultaneous formulation and the separation-aware objective rather than from properties of any single cohort.

Of note, our evaluations are, to our knowledge, the most extensive and heterogeneous reported so far for deep-learning cortical surface reconstruction, spanning 560 subjects across a wide range of ages, clinical populations, and scanner vendors, which is substantially broader than the one or few cohorts typically used to evaluate prior methods. 
We also share all preprocessed data and data splits to allow for reproducibility studies and evaluation of future novel methods.

The ablation study separates the components that drive accuracy from those that maintain geometric validity, and shows that these are not the same components. 
Surface-distance accuracy is governed by the architecture: the residual, group-normalized backbone contributes the largest share, followed by the geometry-encoder depth, the dedicated ribbon encoder with gated fusion, and the ribbon input itself. 
Removing the separation and nesting losses, by contrast, leaves mean accuracy unchanged and instead eliminates the collision result: union collision rises by roughly three orders of magnitude and no case remains collision-free, against $92.14\%$ for the full model. 
The losses therefore act almost entirely on inter-surface validity rather than on surface localization.

The architectural variants also reveal a tradeoff that a collision rate reported in isolation would obscure. 
Because initialization is collision-free in every case, collisions can arise only during deformation, so variants that
fit the target surfaces less closely deform less and collide less: the least accurate variant has the lowest collision rate of any model tested. 
The separation and nesting losses are what allow the full model to occupy the accurate end of this range without paying the corresponding collision cost.

Several limitations should be noted, the first group concerning the reference standard and the evaluation itself. 
The reference surfaces were produced by FreeSurfer and used as a silver standard rather than a true ground truth, so all distance and thickness metrics are defined relative to FreeSurfer and inherit its own inaccuracies. 
Cortical thickness was estimated as a nearest-surface white-to-pial distance and is not based on anatomical vertex correspondence, so the reported thickness error should be read as a geometric agreement measure. 
Some same-hemisphere white--pial contacts near the medial wall may be counted as collisions even though they are anatomically expected, which can slightly inflate collision values, including for the reference pipeline. 

SimCortex v2 used T1w MRI mapped to MNI152 space and did not assess robustness to substantially different acquisition protocols or pathology-related morphology beyond the included cohorts.

Additionally, we made the deliberate choice not to retrain the baseline methods. 
This allows us to compare methods with their best configuration, but is less fair because the training data used may differ not only in terms of dataset size and samples, but also in the labeling of the silver standard as not all methods used FreeSurfer 7.4.1 to generate ground truth surfaces. 
Nevertheless, the primary strength of SimCortex v2 is competitive accuracy combined with geometric validity, and the current evaluation experiment clearly shows the geometric validity advantage of our method.

A second group of limitations concerns the data and the generality of the resulting model. 
The quality of the training data is one: all training scans were high-quality research-grade T1w images, so the model has not been exposed to the lower resolution, reduced contrast, motion, and other artifacts common in routine clinical acquisitions; unlike methods designed for arbitrary clinical data, which are trained across a wide range of acquisition conditions, SimCortex v2 has seen only a narrow, high-quality distribution, and its accuracy is therefore likely to degrade on typical clinical scans. 
SimCortex v2 also does not establish vertex-wise correspondence: each subject's surfaces are initialized from that subject's own segmentation, so the meshes differ in vertex count and connectivity across subjects and within a subject, unlike template-deformation methods, which propagate a single mesh with fixed topology to every subject.
SimCortex v2 would require an additional surface-registration or resampling step before vertex-wise group analyses. Finally, the method was evaluated on geometric metrics; its effect on downstream morphometric and population-level analyses was not measured. 
Future work will establish cross-subject correspondence (for example, via spherical registration or template resampling), evaluate the reconstructed surfaces on regional morphometry and longitudinal analyses, and test generalization to additional acquisition conditions, modalities, and clinical populations.

\section{Conclusion}
\label{sec:conclusion}

We presented SimCortex v2, a framework for simultaneous reconstruction of the left and right WM and pial surfaces from T1w MRI. 
The method combines 9-class anatomical segmentation, hierarchical topology-corrected surface initialization, a cortical ribbon probability map, and joint multi-scale diffeomorphic deformation of all four surfaces, trained with an objective that explicitly regularizes inter-surface separation.

Evaluated on 560 cases from 14 cohorts, SimCortex v2 matched the surface-distance accuracy of the strongest baseline while producing near-zero inter-surface collisions and the lowest self-intersection fraction among the learning-based methods, and it maintained this behavior consistently across cohorts and across all four surfaces. 
Its distinguishing property is thus the combination of accurate surface placement with geometric validity, rather than an improvement in accuracy alone. 
Establishing its value for applied neuroimaging will require extending the framework toward anatomical correspondence and region-specific analyses, which we leave to future work.

\section{Data and software availability}
\label{sec:data_availability}

The processed derivatives generated in this study are available through the Federated Research Data Repository (FRDR) as the \emph{SimCortex Reproducibility Dataset for Cortical Surface Reconstruction}, version 1.0.0 \citep{Moradkhani:2026}, available at
\url{https://doi.org/10.20383/103.01769}. The deposit contains the FreeSurfer 7.4.1 derivatives used in this study for the cases described in Section~\ref{sec:datasets}. For the HCP-YA and OASIS-1 cohorts, it also includes outputs from the SimCortex preprocessing (\texttt{sc-preproc}), nine-class segmentation (\texttt{sc-seg}), and initial-surface generation (\texttt{sc-initsurf}) stages. The exact training, validation, and test splits are provided together with the subject-selection lists used to construct the
external evaluation cohorts. Raw source MRI data are not redistributed and must be obtained from their respective repositories.

Reuse of the deposited derivatives remains subject to the terms associated
with the corresponding source datasets and software. These include the HCP
Open Access Data Use Terms, the OASIS Data Use Agreement, the FreeSurfer
Software License, and the licenses specified by the corresponding OpenNeuro
datasets. The FRDR deposit records the applicable terms for each component;
no repository-level license supersedes those terms.

The SimCortex source code is publicly available at
\url{https://github.com/Neuro-iX/SimCortex}. The experiments reported in this
study correspond to release \texttt{v2.0.0}, available at
\url{https://github.com/Neuro-iX/SimCortex/releases/tag/v2.0.0}. The
repository provides installation instructions, pretrained-model access,
configuration files, and commands for running the complete reconstruction
pipeline.

\printcredits

\section*{Declaration of generative AI and AI-assisted technologies in the manuscript preparation process}

During the preparation of this work, the authors used GPT-5.6 Sol (OpenAI) and Claude Opus 5 (Anthropic) to assist with language editing and with drafting and restructuring portions of the text. After using these tools, the authors reviewed and edited all content, verified all technical claims, results, and references against the underlying data and code, and take full responsibility for the content of the publication.

\section*{Declaration of competing interests}
The authors declare that they have no known competing financial interests or personal relationships that could have appeared to influence the work reported in this paper.

\section*{Acknowledgments}
This research was supported by the Canada Research Chairs Program (CRC-2022-00183) and the Canada Foundation for Innovation. Kaveh Moradkhani was supported by a doctoral research scholarship from the Fonds de recherche du Québec (FRQ), B2X program (award 369771).
\bibliographystyle{cas-model2-names}

\bibliography{references}



\end{document}